\pdfoutput=1

\documentclass[11pt]{article}

\usepackage[review]{acl}

\usepackage{times}
\usepackage{latexsym}
\usepackage{multicol}
\usepackage{multirow}
\usepackage{tabularx}
\usepackage{array}
\usepackage{tabularray}
\usepackage{float}
\usepackage{stfloats}
\usepackage{graphicx}
\usepackage{xtab}
\usepackage{longtable}
\usepackage{xltabular}
\keepXColumns
\usepackage{amsmath}
\usepackage{algorithm,algorithmic}
\usepackage{amsfonts}
\usepackage{xcolor}
\usepackage{lipsum}

\usepackage[T1]{fontenc}

\usepackage[utf8]{inputenc}

\usepackage{microtype}

\title{An Empirical Study of Data Ability Boundary in LLMs' Math Reasoning}
\author{
    Zui Chen$^\Diamond$\textsuperscript{\rm 1,\rm2}, 
    Yezeng Chen$^\Diamond$\textsuperscript{\rm 1,\rm2}, 
    Jiaqi Han\textsuperscript{\rm 1,\rm 2}, 
    Zhijie Huang\textsuperscript{\rm 1,\rm 2}, 
    Ji Qi, 
    Yi Zhou$^\clubsuit$\textsuperscript{\rm 3} \\
    \textsuperscript{\rm 1}School of Information Science and Technology, ShanghaiTech University \\
    \textsuperscript{\rm 2}Shanghai Innovation Center for Processor Technologies \\
    \textsuperscript{\rm 3}School of Information Science and Technology, University of Science and Technology of China  \\
    \texttt{\{chenzui2022, chenyz2022, hanjq2022, huangzhj1\}@shanghaitech.edu.cn;} \\
    \texttt{qiji@cmss.china.mobile.com; }\texttt{yi\_zhou@ustc.edu.cn} 
}

\begin{document}
\maketitle


\begin{abstract}
Large language models (LLMs) are displaying emergent abilities for math reasoning tasks,
and there is a growing attention on enhancing the ability of open-source LLMs through supervised fine-tuning (SFT).
In this paper, we aim to explore a general data strategy for supervised data to help optimize and expand math reasoning ability.
Firstly, we determine the ability boundary of reasoning paths augmentation by identifying these paths' minimal optimal set.
Secondly, we validate that different abilities of the model can be cumulatively enhanced by \textbf{M}ix of \textbf{M}inimal \textbf{O}ptimal \textbf{S}ets of corresponding types of data, 
while our models \textbf{MMOS} achieve SOTA performance on series base models under much lower construction costs.
Besides, we point out GSM-HARD is not really hard and today's LLMs no longer lack numerical robustness.
Also, we provide an Auto Problem Generator for robustness testing and educational applications.
Our code and data are publicly available at \url{https://github.com/cyzhh/MMOS}.
\end{abstract}

\section{Introduction}
In the context of significant emergent abilities demonstrated by Large Language Models (LLMs) (\citealp{weiEmergentAbilitiesLarge2022}; \citealp{openaiGPT4TechnicalReport2023}), the focus on math reasoning tasks, particularly Numerical QA and Math Word Problems (MWP) (\citealp{kushmanLearningAutomaticallySolve2014}; 
\citealp{upadhyayAnnotatingDerivationsNew2017}; 
\citealp{miaoDiverseCorpusEvaluating2020a}; \citealp{xuRobustNumericalQuestion}), is paramount.
The current approach to activate these abilities in LLMs involves carefully engineered prompting \citep{brownLanguageModelsAre2020b}, in-context learning (ICL) \citep{chenMetalearningLanguageModel2022} or supervised fine-tuning (SFT). 

Particularly due to computational costs and stability concerns \citep{yuanRFT2023}, there is growing attention on enhancing the abilities of open-source LLMs \citep{roziereCodeLLaMAOpen2023} through SFT. 
Supervised data is crucial for SFT. 
Current research is centered on using GPT-4 \citep{openaiGPT4TechnicalReport2023} or other powerful base models with prompts composed of their designed reasoning chains to create supervised data for SFT based on several public seed datasets \citep{luSurvey2022}.

In this paper, we aim to explore a general data strategy for supervised data to help optimize and expand math reasoning ability. We primarily investigate the following research questions (RQs):

\begin{itemize}
  \item[\textbullet] RQ1: 
What is the ability boundary of reasoning paths, and how to select paths optimally?
  \item[\textbullet] RQ2: 
  How can we expand the ability boundary, and what kinds of problem sets are needed for this expansion?
\end{itemize}

\begin{figure}[t]
  \centering
  \includegraphics[width=\linewidth]{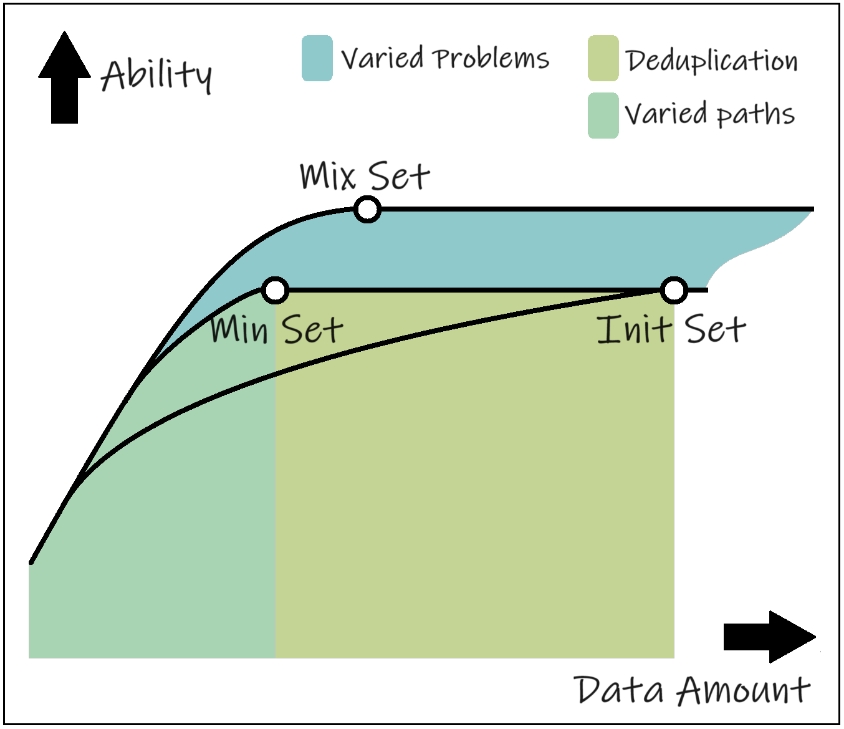}
  \caption{Conceptual figure of the ability boundary}
  \label{fig:first_table}
\end{figure}

RQ1 originates from a common challenge in response augmentation methods: determining the optimal amount of data the training set should cover to balance data amount, effectiveness, and generalizability.
As for RQ2, we focus on introducing additional problems instead of synthesizing new questions for query augmentation, which could assist in selecting and combining the necessary data from the chaotic reality of existing datasets.
Actually, we first explore methods to enhance weak ability, and then focus specifically on Out-Of-Domain (OOD) ability, numerical robustness, and further extending the model’s existing ability.

The overall data strategy is illustrated in Figure \ref{fig:first_table}. Based on the initial set obtained from n-sampling, we determine the ability boundary of reasoning paths augmentation and then achieve optimization by identifying the Minimal Optimal Set (MOS) for individual datasets through deduplication. Furthermore, we facilitate expansion by creating \textbf{M}ix of \textbf{M}inimal \textbf{O}ptimal \textbf{S}ets (MMOS).

The findings for RQ1 (1,2) and RQ2 (3,4,5) include the following points:

1. Providing varied, deduplicated and correct reasoning paths can improve math reasoning ability in In-Domain and Similar-Domain data. (Sec \ref{subsec:3.3})

2. The ability boundary of increasing reasoning paths is reached, that is, we identify the minimal optimal set, when the number of paths is similar to the number of distinct problem solutions. (Sec \ref{subsec:3.4})

3. Different abilities of the model can be cumulatively enhanced by mixing minimal optimal sets of corresponding types of data. (Sec \ref{subsec:4.2})

4. GSM-HARD is not really hard and the numerical robustness issue is no longer prevalent in today's LLMs. We also build a high-quality Auto Problem Generator for these numerical robustness tests and educational applications. (Sec \ref{subsec:4.3} \& \ref{subsec:4.4})

5. An overlapping dataset can continue to enhance the model's ability in the absence of corresponding data. And MMOS which has much lower construction costs can also achieve SOTA performance on series base models. (Sec \ref{subsec:4.5} \& \ref{subsec:4.6})

\section{Ability Boundary of Reasoning Paths}
\subsection{Overview}
In this section, for RQ1, we aim to determine the ability boundary of reasoning paths and find a data strategy.
We hypothesize that a minimal set capable of maximizing math reasoning ability consists of varied, deduplicated and correct reasoning paths. 

In following Section \ref{subsec:3.2}, we discuss about the datasets. 
In Section \ref{subsec:3.3}, we identify this minimal optimal set and determine the benefits of removing duplicates and keeping varied reasoning paths within a certain range.
In Section \ref{subsec:3.4}, we employ a clustering method as a filter to further explore the boundary. 
In Section \ref{subsec:3.5}, we conduct an ablation experiment to assess the impact of ensuring the correctness of the reasoning paths.
 
All detailed experiment settings are in \ref{subsec: Experiment Setup}.

\subsection{Dataset Comparation}
\label{subsec:3.2}


Six datasets are involved in this study. Detailed information about their origins, example analyses, and a preliminary estimation of their difficulty levels can be found in the Appendix \ref{sec:Datasets}.

\begin{figure}[h]
  \includegraphics[width=\linewidth]{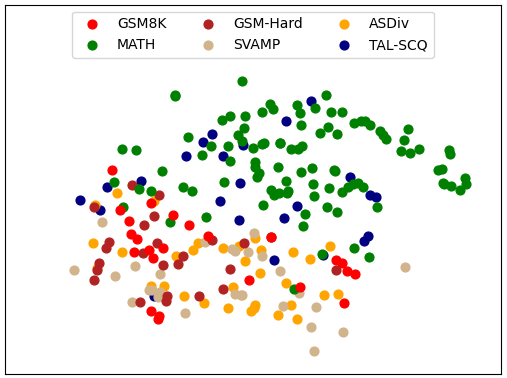}
  \caption{Visualization of query embedding distribution 
 through t-SNE across six distinct datasets.}
  \label{fig:data_emb}
\end{figure}

To better understand the problems' difference across these datasets, we visualize the hidden representations of problems using t-SNE. 
This visualization as Figure \ref{fig:data_emb} reveals a notable separation in the distribution of problems from the GSM8K and MATH datasets into two distinct clusters. 
This divergence emphasizes the contrast in question styles: GSM8K being text-intensive, while MATH is more focused on math expressions.

For the experiments presented in this section, we exclusively use GSM8K without bootstrapping its questions. 
Consequently, GSM8K is categorized as our IND data. Conversely, the MATH dataset, with its significant stylistic and content differences, is classified as OOD data.
Additionally, two other datasets, SVAMP and ASDiV, although different in origin from GSM8K, show similarities in both question types and spatial representations. 
Therefore, we consider these to be Similar-Domain Datasets. And we denote SVAMP and ASDiV as S\&A in the subsequent analysis.


\subsection{Identify the Minimal Optimal Set} 
\label{subsec:3.3} 

To identify the minimal optimal set, we follow these steps:
1) Sample a sufficient number of correct reasoning paths to form initial set.  
2) Implement a deduplication algorithm to obtain its deduplicated subset.
3) Conduct a statistical analysis on the upper limit of reasoning paths per question k with the subset data amount N.
4) Perform SFT on several subsets to analyze the impact of removing duplicates and keeping varied reasoning paths.


\textbf{Initial set} created by various original methods face API and learning costs, and there is a scarcity of training data being open-sourced. Therefore, we attempt to directly use open-source models. Specifically, we opt for advanced models ToRA \citep{gouToRAToolIntegratedReasoning2023} that combine programs and rationales, and apply rejection sampling \citep{yuanRFT2023} to build initial set. And this method, resembling self-learning, possesses a certain degree of universality.

We employ four pre-trained models: ToRA-CODE 7B/13B/34B and ToRA 70B. 
For every question in the GSM8K dataset, these models sample 100 reasoning paths each with temperature 0.9. 
We then merge 400 reasoning paths and extract those whose code can be executed and have correct answers to obtain the initial training set $E_{u400}$.

\begin{figure*}[t]
  \centering
  \includegraphics[width=1\textwidth]{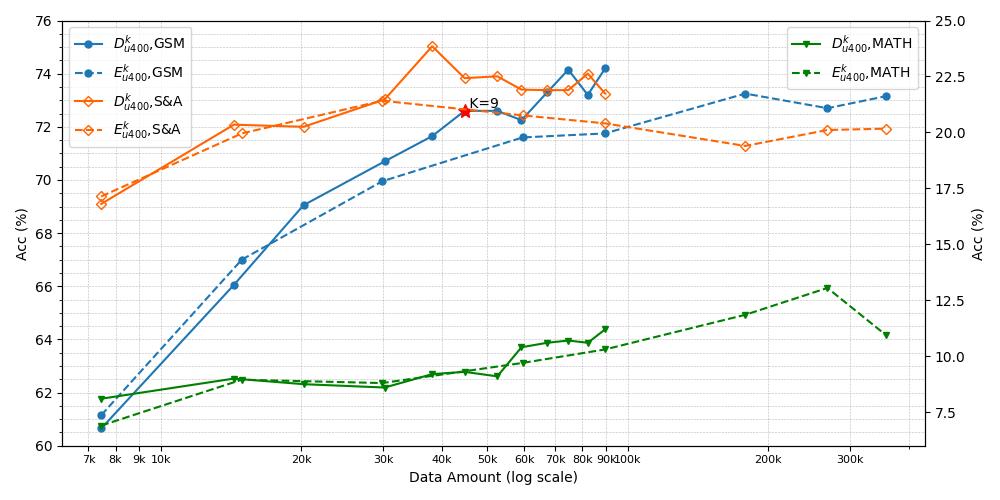}

      \caption{Comparison of test set accuracy on GSM8K, S\&A and MATH for models after SFT on Code LLaMA 7B using series subsets of $D^{k}_{u400}$ and $E^{k}_{u400}$ with different data amount.}
  \label{fig:nodup_vs_dup}
\end{figure*}



\begin{algorithm}[t]
\caption{Deduplicate Data by Codes}
\begin{algorithmic}[1]
\label{algorithm:1}
\REQUIRE data d, extract $\xi(\cdot)$, recovery $\widetilde{\xi}(\cdot)$, astparse $\mathbb{P}(\cdot)$, astunparse $\widetilde{\mathbb{P}}(\cdot)$, deduplicate $\widetilde{\mathbb{D}}(\cdot)$
\FOR{$i = 1$ \TO $n$}
    \STATE $c_i \leftarrow \xi(d_i | q_i \oplus a_i \oplus s_i)$ \hfill $\triangleright$ Code Extraction
    \STATE $t_i \sim \mathbb{P}(c_i)$ \hfill $\triangleright$ Code Astparse
    \STATE $t^{'}_{i} \leftarrow \pi(t_i | v \oplus f)$ \hfill $\triangleright$ Code Substitution
    \STATE $c^{'}_i \sim \widetilde{\mathbb{P}}(t^{'}_{i})$ \hfill $\triangleright$ Code Astunparse
\ENDFOR
\STATE $c^{'} \leftarrow \mathbb{D}(c^{'})$ \hfill $\triangleright$ Code Deduplication
\STATE $d^{'} \leftarrow \widetilde{\xi}(c^{'} \oplus q \oplus a \oplus s)$ \hfill $\triangleright$ Data Recovery
\end{algorithmic}
\end{algorithm}

\textbf{Deduplication Algorithm} \ref{algorithm:1} aim to extract the deduplicated subset $D_{u400}$ from $E_{u400}$ by codes which share the same calculation process.

We iterate all n data with following steps:

1) Extract the code block $c_i$ from data $d_i$, which includes query $q_i$, completion $a_i$ and source $s_i$ .

2) Employ the Abstract Syntax Tree (AST) method to parse the code into the tree $t_i$.

3) Normalize the tree by replacing variable names 
$v$ with lowercase letters and function names 
$f$ with uppercase letters, resulting in $t{'}_i$.

4) Convert the normalized tree back into normalized code, denoted as $c^{'}_i$.

After completing the iteration, the normalized codes are duplicated through plain text matching. 
Finally deduplicated data $d^{'}$ is recovered with the deduplicated code, query, completion and source.


\textbf{The k-N relation}
can be regarded as an estimation of the relationship between the number of reasoning paths per question k and the corresponding subset data amount N. This relation is obtained by implementing an upper limit on the reasoning paths per question in the initial set.

As shown in Appendix \ref{appendix:relation_of_k&N}, the k-N curve demonstrates a linear relationship 
on $E_{u400}$ with a median of $k=400$ and a mean of $k=392.14$. 
In contrast, on $D_{u400}$, it exhibits a log-linear relationship with a median of 7 and a mean of 12.01. This indicates that the deduplication method is effective but still leaves room for improvement.

\begin{table*}\small
\centering
{%
\begin{tabular}{|c|c|cccccc|}
\hline


& k & 5 & 7 & 9 & 15 & 27 & - \\
\cline{2-8}
$D^{cluster,k}_{u400}$ & GSM8K & 71.4(\textcolor{red}{+0.7}) & 70.9(\textcolor{green}{-0.7}) & 72.6(\textcolor{green}{-1.2}) & 73.4(\textcolor{red}{+0.8}) & 74(\textcolor{green}{-0.1}) & -\\
& S\&A & 73.4(\textcolor{red}{+0.5})   & 73.4(\textcolor{green}{-0.9})  & 73.1(\textcolor{green}{-0.9})  & 74.2(\textcolor{red}{+0.6}) & 73.4(\textcolor{red}{+0.0}) &  - \\
\hline
&  k & 2 & 4 & 8 & 12 & 24 & 36  \\
\cline{2-8}
$E^{cluster,k}_{u400}$ & GSM8K & 67.6(\textcolor{red}{+0.6}) & 70.5(\textcolor{red}{+0.6}) & 72.1(\textcolor{red}{+0.5}) & 74.0(\textcolor{red}{+2.3})& 73.2(\textcolor{red}{+0.0})  & 73.5(\textcolor{red}{+0.8}) \\
& S\&A  & 72.0(\textcolor{red}{+0.3})  & 71.8(\textcolor{green}{-1.1})  & 74.4(\textcolor{red}{+2.0}) & 72.3(\textcolor{red}{+0.2}) & 73.0(\textcolor{red}{+2.0}) & 73.3(\textcolor{red}{+1.4})  \\

\hline
\end{tabular}
}
      \caption{Comparison of test set accuracy on GSM8K and S\&A for models after SFT on Code LLaMA 7B using series subsets of $D^{k}_{u400}$ and $E^{k}_{u400}$ through clustering.}
\label{tab:cluster}
\end{table*}
\begin{table}
\centering
\small
{%
\begin{tabular}{ccccc}
\hline
\textbf{Dataset} & \textbf{k} & \textbf{N} & \textbf{GSM8K} & \textbf{S\&A}\\
\hline
$D^{k}_{u400}$ & 9 & 44771 & 71.4 & 73.6 \\
$D^{total,k}_{u400}$ & 9 & 46740 & 69.8(\textcolor{green}{-1.6}) & 73.7(\textcolor{red}{+0.1}) \\
$D^{k}_{u400}$ & $\infty$ & 89530 & 74.2 & 73.3 \\
$D^{total,k}_{u400}$ & $\infty$ & 126391 & 71.7(\textcolor{green}{-2.5}) & 73.0(\textcolor{green}{-0.3}) \\
\hline
\end{tabular}
}
      \caption{Comparison of test set accuracy on GSM8K and S\&A for models after SFT on Code LLaMA 7B using $D_{u400}$ and $D^{total}_{u400}$.}
\label{tab:true_vs_all}
\end{table}
\textbf{Comparative experiment} includes two aspects. Firstly, to verify the effectiveness of adding varied paths, we conduct random selection of k paths for each question within $D_{u400}$ to obtain twelve $D^k_{u400}$ subsets with 
k $\in$ \{1,2,3,5,7,9,12,15,20,27,40,$\infty$\}, 
N $\in$ \{7.5,15,20,30,38,45,53,60,67,75,82,90\}K.

Secondly, to better assess the impact of duplicate removal, we maintain a consistent order of magnitude in terms of data amount on $E_{u400}$ and obtain $E^k_{u400}$ with k$\in$\{1,2,4,8,12,24,36,48\} and N$\in$\{7.5,15,30,60,90,180,270,360\}K.


\textbf{Evaluation \& Conclusion.} 
We conduct SFT on Code LLaMA 7B using a series of subsets $D^k_{u400}$ and $E^k_{u400}$, and then inference on the test split of GSM8K, S\&A, and MATH.  

Results are shown in Figure \ref{fig:nodup_vs_dup}. 
On the IND dataset GSM8K, as indicated by the blue solid line, the model's ability maintains a linear relationship with the logarithm of data amount before $k=9$, $N=45K$.
In contrast, the blue dashed line representing the initial set data aligns with this trend only when k is small and duplicate paths are less likely to be selected. Beyond this point, further increasing the data amount sharply diminishes the marginal improvement in model ability. 
This suggests that enhancing the model's ability stems from adding varied reasoning paths, rather than merely increasing the data amount.

We also observe that with the same data amount, beyond $N = 30K$, the performance on $D_{u400}$ consistently surpasses that on $E_{u400}$. 
This reflects that removing duplicates can not only diminish the training duration but also enhance the model's ability.

On the Similar-Domain Datasets S\&A, potentially due to the inherently easier nature of the questions, the models achieve high effectiveness even at k=1. 
The other conclusions are similar to those observed on GSM8K.

However, on the OOD dataset MATH, the models consistently exhibit weaker ability. This may be, as shown in Section \ref{subsec:3.2}, due to the differing types of questions presented in the dataset.

Thus far, we have essentially reached the conclusion that providing varied, deduplicated, and correct reasoning paths can improve math reasoning ability in both IND and Similar-Domain data.

Finally, we conduct a case study, as shown in Appendix \ref{appendix:minimal_sets}, where our example problem has 10 different solutions which is similar to the previously inflection point of k=9.
Therefore, we consider $D^{k=9}_{u400}$ as the minimal optimal set. 
From this, we draw another conclusion: the ability boundary is reached, that is, we identify the minimal optimal set, when the number of reasoning paths is similar to the number of potential problem solutions.

\subsection{Cluster as a Filter}
\label{subsec:3.4}

Our deduplication algorithm, as an extension of a template method, is not flawless and can fail to eliminate similar paths. 
The example problem shown in Appendix \ref{appendix:minimal_sets}
has only 10 distinct solutions. However, in $D_{u400}$, 43 paths are still retained. When we implement random selection to obtain $D^{k=9}_{u400}$, it only includes 6 distinct solutions.


We attempt to use clustering as a filter, replacing random selection, in order to ensure that the resulting $D^{k=9}_{u400}$ subset contains a greater number of distinct solutions.
Specifically, we first obtain the embedding vectors of the codes. Then, we apply Latent Semantic Analysis (LSA) for dimensionality reduction, followed by k-means clustering. We extract and retain the central data points from these clusters. On the same example problem, the new $D^{k=9}_{u400}$ contains 7 distinct solutions.



In the comparative experiment, we replace random selection with clustering to obtain new subsets, $D^{cluster,k}_{u400}$ and $E^{cluster,k}_{u400}$. We then conduct SFT on Code LLaMA 7B using these subsets.


As shown in Table \ref{tab:cluster}, the results on $E^{cluster,k}_{u400}$ exhibited a consistent improvement, suggesting that using clustering as a filter is viable. However, this is not the case for $D^{cluster,k}_{u400}$. We speculate that the remaining similar paths after deduplication have only a minor impact.

\subsection{Correct Reasoning Ablation}
\label{subsec:3.5}

While ensuring the correctness of paths is intuitively sound, we also observe that some methods, despite not guaranteeing correct answers for created problems, still yield reasonably good results. Therefore, we aim to ablate the effect of ensuring the correctness of paths.


During the acquisition of the initial set $E_{u400}$, we retain all data, including those with incorrect answers, resulting in $E^{total,k}_{u400}$. After deduplicating this set, we obtain $D^{total,k}_{u400}$.
Subsequently, we generated subsets for $k=9$ and $k=\infty$ through random selection from these sets and conduct comparative experiments with these subsets.



As illustrated in Table \ref{tab:true_vs_all}, on GSM8K, not filtering out incorrect paths leads to a noticeable decline in performance. However, this effect is not observed on S\&A, which could be attributed to the lower difficulty level of S\&A.

\begin{figure*}[t]
  \centering
  \includegraphics[width=1\textwidth]{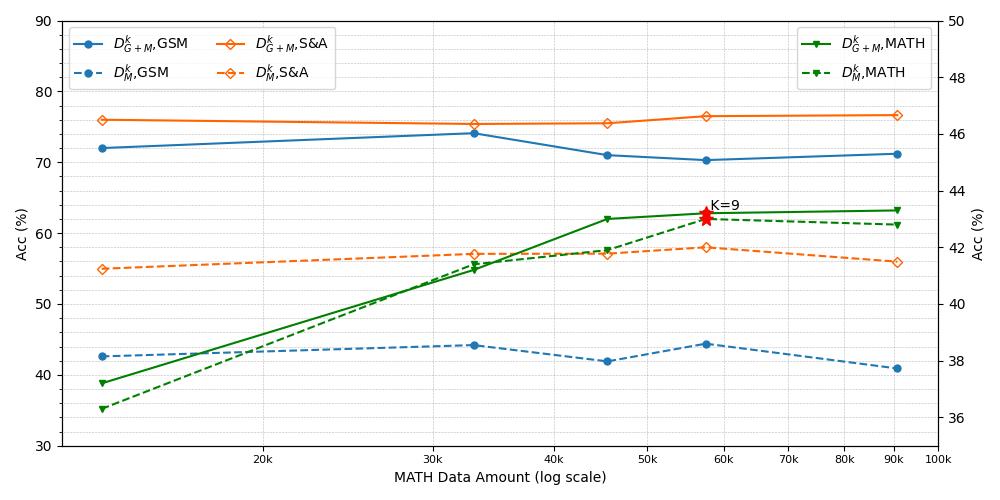}
    \caption{Comparison of test set accuracy on GSM8K, S\&A and MATH for models after SFT on Code LLaMA 7B using series subsets of $D^{k}_{G+M}$ and $D^{k}_{M}$ with different MATH data amount.}
    \label{fig:weak_ability}
\end{figure*}

\section{Expand Boundary with Problems}
\subsection{Overview}
In this section, for RQ2, we consider expanding this ability boundary by introducing additional problems. 
We first explore methods to enhance weak ability, and then focus specifically on OOD ability, numerical robustness, and further extending the model’s existing ability.

In Section \ref{subsec:4.2}, we examine whether the model's weak ability can be enhanced by providing corresponding data.
Section \ref{subsec:4.3} delves into the robustness of the model's numerical abilities and the issues present in a dataset, GSM-HARD.
In Section \ref{subsec:4.4}, we develop an automated, high-accuracy problem generator for constructing numerically perturbed data, demonstrating its practical application value. 
Finally, in Section \ref{subsec:4.5}, we strive to achieve a state-of-the-art model and discuss the potential for further extending the model's existing ability.


\subsection{Enhance Weak Ability}
\label{subsec:4.2}

To address the issue of the weak ability of models trained with the minimal optimal set of GSM8K when applied to OOD set MATH, a straightforward solution is to provide corresponding data.


Initially, following the same method described in Section \ref{subsec:3.2}, we obtain a series of deduplicated subsets $D^{k}_{M}$ constructed using the MATH dataset and subsequently conduct SFT on them.
And, as indicated by the green dashed line in Figure \ref{fig:weak_ability}, we identify the minimal optimal set $D^{k=9}_{M}$ on MATH.
As expected, compared to the models trained on $D^{k}_{u400}$ originating from GSM8K, there is a significant improvement in ability on MATH, and the abilities on GSM8K and S\&A, represented by the blue and yellow dashed lines, are weaker.


Subsequently, we merge the subsets $D^{k}_{M}$ from the MATH dataset with the minimal optimal set of GSM8K $D^{k=9}_{u400}$, denoted as $D^{k}_{G+M}$.
The experimental results on $D^{k}_{G+M}$, as shown by the various solid lines, indicate that compared to $D^{k}_{M}$, which provides the same amount of data from MATH, there is a slight improvement in performance on MATH, and a significant improvement on GSM8K and S\&A. 
Additionally, the local optimum point for $D^{k}_{G+M}$, similar to $D^{k}_{M}$, is also achieved at k=9.
Similarly, compared to $D^{k=9}_{u400}$, $D^{k=9}_{G+M}$ shows a slight decrease in performance on GSM8K, dropping from 72.6\% to 70.3\%, and a marginal decline on S\&A, going from 73.6\% to 76.5\%. However, there is a significant improvement on MATH, with a rise from 10.4\% to 43.2\%.
Overall, $D^{k=9}_{G+M}$ (102K) effectively combines the strengths of $D^{k=9}_{u400}$ (45K) and $D^{k=9}_{M}$ (57K), showcasing enhanced abilities on GSM8K, S\&A, and MATH datasets.

We arrive at a fundamental conclusion: different abilities of the model can be cumulatively enhanced by mixing minimal optimal sets of corresponding types of data. 
This finding provides a simple yet effective method for enhancing the model's weak abilities by acquiring the corresponding datasets.



\subsection{Is GSM-HARD Really Hard?}
\label{subsec:4.3}



Another 'weak ability' of the $D^{k=9}_{G+M}$ model is demonstrated on GSM-HARD (54.8\% vs 70.3\% on GSM8K). This dataset is created by replacing the numbers in GSM8K with larger ones \citep{gaoPALProgramaidedLanguage2023}. Given that only the numerical values are altered, the distribution of problems in Figure \ref{fig:data_emb} remains almost identical. Based on the conclusions from Section \ref{subsec:3.3}, such a significant discrepancy should not occur, whether we consider it as IND or Similar-Domain data.
This leads us to questions: Is GSM-HARD really hard? Is the model's numerical robustness indeed weak?



The first source of discrepancy arises from the standards of ground truth. Due to the lack of meticulous design in the numerical values of the questions, some answers are not impractical, such as receiving answers with decimals when asking about quantities, or negative numbers when asking about the amount decreased. In practice, these initial calculation results should be rounded or converted to absolute values when providing answers, but GSM-HARD directly annotates these initial calculation results as the ground truth.
We do not consider this to be indicative of a gap in ability. Therefore, using the standards of GSM-HARD, and evaluating based on the initial calculation results, the accuracy rate increases to 63.3(\textcolor{red}{+8.5})\%.

The second source of the discrepancy is due to errors in the ground truth annotation, stemming from an imperfect automated annotation process in GSM-HARD after modifying the problems. The corresponding values in the code are not updated in line with the changes in the numerical values in the problems, thus leading to execution with retained incorrect results as ground truth.
We review the first 50 samples where the $D^{k=9}_{G+M}$ model make incorrect inferences and discover 25 errors in the ground truth annotations. We can estimate that the remaining gap, 
70.3\% - 63.3\% = 7\% < (1 - 63.3\%) * (25/50) * 63.3\% 
can be covered by these annotation errors.
Finally, we conjecture that GSM-HARD is not really hard and the numerical robustness issue is no longer prevalent in today's LLMs.

\subsection{Auto Problem Generator}
\label{subsec:4.4}

Considering this, developing an Auto Problem Generator capable of reliably producing data similar to GSM-HARD is meaningful. Such a generator can be used to test the numerical robustness of models. Additionally, it can also be utilized in educational applications to assess students' abilities.


Auto Problem Generator follows these steps:

1) Generate the deduplicated subset $D_{test,u400}$ from the seed dataset, the test split of GSM8K, following the method in Section \ref{subsec:3.3}. 

2) For each question, extract the reasoning path with the highest repetition as the main path and separate the remaining path as the remain paths.

3) Extract numbers from questions using template matching and modify them with function $f(\cdot)$.


4) Modify the corresponding numbers in the code of the main path and execute it to obtain the answer $A_{main}$. 

5) If the code execution fails or $A_{main} < 0$, modify the numbers again with 50 times limit.

6) Repeat step 4 on the remaining paths and obtain the answer set $A_{remains}$.

7) If all elements in $A_{remains}$ are identical to $A_{main}$, then we believe $A_{main}$ is correct.

8) Combine the correct $A_{main}$ with the modified questions to form the generated dataset P.


We apply the Distribution Perturbation \citep{xuRobustNumericalQuestion} on numerical values with the following function $f(n)$ with $\mu$=5, $\sigma$=1 and $\mu$=1000, $\sigma$=300 to create datasets $P_{5}$ and $P_{1000}$,
\[
    f(n) = n + \lfloor X \rfloor, X \sim \mathcal{N}(\mu, \sigma^2)
\]
that $\mathcal{N}$ represents normal distribution.
We manually review the first 100 QA pairs in $P_{5}$ and achieve a 98\% accuracy rate, with only two questions having incorrectly annotated answers. A detailed analysis of these errors and their reasons can be found in the Appendix \ref{sec:Error Analysis}.

Thus, we have successfully developed a high-quality Auto Problem Generator, which can be used for testing the numerical robustness of models as well as for educational application.


\begin{table*}[ht]\small
  \centering
\begin{tabular}{c|c c c c|c c c c}
\hline 
\multirow{2}{*}{\textbf{Model}} & \textbf{GSM8K} & \textbf{SVAMP} & \textbf{ASDiv} & \textbf{MATH} & \textbf{GSM8K} & \textbf{SVAMP} & \textbf{ASDiv} & \textbf{MATH} \\
\cline{2-9}
& \multicolumn{4}{c|}{7B} & \multicolumn{4}{c}{13B} \\
\hline
LLaMA-2 & 13.3 & 38.0 & 50.7 & 4.1 & 24.3 & 43.1 & 56.3 & 6.3 \\
LLaMA-2 SFT & 41.3 & 31.9 & 47.4 & 7.2 & 51.1 & 46.3 & 58.6 & 9.2 \\
LLaMA-2 RFT & 50.3 & - & - & - & 55.4 & - & - & -  \\
WizardMath & 54.9 & 57.3 & 59.1 & 10.7 & 63.9 & 64.3 & 65.8 & 14.0 \\
MAmmoTH & 53.6 & 67.7 & - & 31.5 & 62.0 & 72.4 & - & 34.2 \\ 
MetaMath & 66.5 & - & - & 19.8 & 72.3 & - & - & 22.4 \\
MathCoder-L & 64.2 & 71.5 & - & 23.3 & 72.6 & 76.9 & - & 29.9 \\
MathCoder-CL & 67.8 & 70.7 & - & 30.2 & 74.1 & 78.0 & - & 35.9 \\
ToRA & 68.8 & 68.2 & 73.9 & 40.1 & 72.7 & 72.9 & 77.2 & 43.0 \\
TORA-CODE & 72.6 & 70.4 & \textbf{78.7} & \textbf{44.6} & 75.8  & 75.7 & 81.4 & 48.1 \\
MMOS & 69.9 & 73.4 & 76.8 & 40.2 & 74.8 & 77.0 & 80.0 & 43.2 \\
MMOS-CODE & \textbf{73.9} & \textbf{76.4} & 78.6 & 44.3 & \textbf{77.1} & \textbf{77.5} & \textbf{81.9} & \textbf{48.1} \\
MMOS-Min-CODE & 70.3 & 72.5 & 76.7 & 44.6  & - & - & - & - \\
\hline 						
\end{tabular}
\caption{Comparison of test set accuracy on 4 datasets for LLaMA-2 and Code LLaMA 7B/13B based models.}
\label{tab:SOTA}
\end{table*}

\textbf{Numerical Robustness} represents a model's consistent ability to handle different types of numerical values. Distribution Perturbation, as applied in GSM-HARD, $P_{5}$, and $P_{1000}$, is one such example.
We evaluate $P_{5}$ and $P_{1000}$ with the model trained on $D^{k=9}_{u400}$ with only GSM8K data.
The experimental results show 73.8\% on GSM8K, 72.1(\textcolor{green}{-1.7})\% on $P_{5}$ and 70.1(\textcolor{green}{-3.7})\% on $P_{1000}$.

Then, employing the same approach, we produce $P^{'}_{1000}$ using the train split of GSM8K and include it in our training data. However, the results show tiny improvement, achieving 73.2\% on GSM8K, 72.6(\textcolor{green}{-0.6})\% on $P_{5}$ and 70.4(\textcolor{green}{-2.8})\% on $P_{1000}$. 
Considering the results of both sets of experiments, since providing corresponding data does not enhance ability, we infer that the discrepancies in $P_{1000}$ are more likely due to annotation issues caused by the inclusion of large numbers.

We also experiment with other numerical perturbation approaches including Language Perturbation and Noise Perturbation.
Language Perturbation does not entail changes to the answers and simply involves converting numerical values into their English word representations. This has led to a slight improvement in the model's performance.
Noise Perturbation introduces noise by adding decimal parts to the numerical values. The conclusions drawn from this method are similar to those from Distribution Perturbation.
Overall, we conclude that current LLMs no longer face significant issues with numerical robustness.

\subsection{Expand Existing Ability}
\label{subsec:4.5}


After utilizing all data from GSM8K and MATH, we try to further expand existing ability in the absence of corresponding data,
As shown in Figure \ref{fig:data_emb}, dataset TAL-SCQ displays query embeddings that overlap with GSM8K and MATH. We generate its minimal optimal set and merge it with $D^{k=9}_{G+M}$, denoted as $D_{G+M+T}$.
Similarly, we conduct SFT on Code LLaMA 7B and achieve an accuracy of 73.9(\textcolor{red}{+3.6})\% on GSM8K, 77.5(\textcolor{red}{+1.0})\% on S\&A, and 44.3(\textcolor{red}{+1.1})\% on MATH.
We conclude that an overlapping dataset can continue to enhance the model's existing ability in the absence of corresponding data.

\subsection{MMOS' Advantage}
\label{subsec:4.6}
Our data strategy MMOS' advantage stands for two aspects, higher performance and lower construction costs.
1) The results, as shown in Table \ref{tab:SOTA}, indicate that our model using MMOS $D_{G+M+T}$ achieves most SOTA performance. 2) 
When constructing the initial set, n-sampling on GPT-4 is costly. Sampling 20 reasoning paths for each seed question of $D_{G+M+T}$ will exceed a cost of \$10,000, and additional learning costs are required for post-processing using various methods. In contrast, MMOS can directly utilize corresponding method models for sampling, avoiding these issues. The sampled data will possess higher quality and lower diversity.
Furthermore, we also attempt to significantly reduce computational costs by sampling 100 solutions for 19k seed questions using only a 7B model. This can be completed within 12 hours on 8 A100 40G GPUs. This approach yields about 30\% amount of the GSM8K reasoning paths and 90\% for MATH, possibly because simpler problems are more prone to repetition. The model obtained, MMOS-Min-CODE, also demonstrates satisfactory performance.

\section{Related Work}
\subsection{LLM for Math Reasoning}

\textbf{Prompt based methods} activate the emergent abilities without training.
A significant breakthrough comes from Chain-of-thought prompting (CoT) \citep{weiChainofThoughtPromptingElicits}, which enhances the ability of LLMs to tackle complex reasoning by using explicit reasoning steps.
The least-to-most prompting strategy \citep{zhouLeasttoMostPromptingEnables2023b} deconstructs complex problems into a series of simpler sub-problems, which are then solved sequentially.
Program of thoughts prompting \citep{chenPoT2022} and program-aided language models \citep{gaoPALProgramaidedLanguage2023} address the limited numerical abilities of LLMs and utilize LLMs solely for understanding problems and generating programs, while offloading computation to an external Python interpreter.

\textbf{Decoding related methods} focus on enhancing performance by replacing the greedy decoding strategy during the inference stage. \citep{wangSelfConsistency2023} samples a diverse set of reasoning paths and selects the most consistent answer, while \citep{xieSelfevaluationGuidedBeam2023} proposes a decoding algorithm that integrates self-evaluation guidance through the use of stochastic beam search.

\textbf{Supervised Fine-tuning (SFT)} based methods are designed to enhance the math reasoning abilities of open-source models such as LLaMA \citep{touvronLLaMAOpenEfficient2023}, LLaMA2 \citep{touvronLLaMAOpenFoundation2023}, and Code LLaMA \citep{roziereCodeLLaMAOpen2023}, while ensuring transparency. Current methods (\citealp{yuMetaMathBootstrapYour2023}; \citealp{wangMathCoderSeamlessCode2023}) largely utilize various prompt-based approaches, employing GPT-4 \citep{openaiGPT4TechnicalReport2023} or other open-source models, to generate reasoning steps as training datasets based on original QA in various datasets like
GSM8k \citep{cobbeTrainingVerifiersSolve2021}, 
MATH \citep{hendrycksMeasuringMathematicalProblem2021}.
These generated reasoning steps can either be in natural language (rationales) \citep{zelikmanSTaRSelfTaughtReasoner2022} or a combination with program (\citealp{yueMAmmoTHBuildingMath2023}; \citealp{gouToRAToolIntegratedReasoning2023}).

\subsection{Supervised Data Augmentation}
\textbf{Response augmentation} approaches (\citealp{luoWizardMath2023}; \citealp{gouToRAToolIntegratedReasoning2023}) involve employing techniques such as nucleus sampling (top-p sampling) \citep{holtzmanCURIOUSCASENEURAL2020} and combining inferences from models of varying sizes, with the aim of enlarging the amount of generated reasoning steps \citep{zhuSolvingMathWord2023}.
These methods generally adhere to an intuitive understanding \citep{niLEARNINGMATHREASONING2022} that fine-tuned models are prone to biases towards a limited set of reference solutions.


\textbf{Query augmentation} methods focus on modifying existing questions to generate new ones. \citet{liDataAugmentation2023} finds that the diversity and complexity of problems contribute positively to performance, and \citet{yuMetaMathBootstrapYour2023} believes that bootstrapping questions can provide multiple perspectives of meta-knowledge, crucial for covering more unseen scenarios and enabling stronger generalization. 
Earlier researches applied Named Entity Recognition or Regular Expression matching to build templates for augmenting questions \citep{liSurveyDeepLearning2022}. \citet{xuRobustNumericalQuestion} focused on categorizing questions based on numerical abilities and designing numerical perturbations. 



\section{Conclusion}



We explore a general data strategy for supervised data to help optimize and expand math reasoning ability.
Firstly, we ascertain the ability boundary related to the augmentation of reasoning paths by identifying the minimal optimal set of these paths, with a focus on maximizing the data's potential.
Secondly, we corroborate the premise that different abilities of the model can be collectively enhanced by amalgamating minimal optimal sets of data, each corresponding to specific types of information. 
Our models achieve SOTA performance on series base models with much lower construction costs.
Additionally, we uncover that LLMs currently do not exhibit a significant lack of numerical robustness.
Moreover, we introduce an Auto Problem Generator, designed for testing the robustness of models and for use in educational applications.






\newpage
\section*{Limitations}



The limitations of our paper include the following aspects:

\textbf{Datasets and Models}. In our research, we use only three datasets to create a mix of minimal optimal sets as training data. However, we are uncertain whether the two conclusions drawn in Section 4 – that different abilities of the model can be cumulatively enhanced by mixing minimal optimal sets of corresponding types of data, and that an overlapping dataset can continue to enhance the model's ability in the absence of corresponding data – would still hold true with the introduction of more and larger datasets. Additionally, we are also unsure if these conclusions would apply to larger-scale models, such as the 70B model.

\textbf{Sampling Bias}. Our conclusions regarding the numerical robustness of the model, the GSM-HARD dataset and the Auto Problem Generator are based on our numerical analysis of accuracy and results from sample checks. This approach may introduce bias.




\section*{Ethical Statements}
 We claim from various aspects that our work is free of ethical risks: 
 
 1) Our research utilizes open-source models like LLaMA-2 and Code LLaMA and open datasets, and we strictly adhere to their licensing protocols. 
 
 2) Despite providing a new auto problem generator,  its functionality is confined to numerical perturbation derived from open-source datasets. We endeavour to prevent the generation of illogical problems and the dissemination of inappropriate information resulting from numerical perturbations. 

 3) During the writing process, we used GPT4 to translate and correct grammatical errors, and the text was human-checked and rewritten to ensure that there were no ethical issues.
 
 4) Our experiments are designed to be resource-efficient, requiring minimal compute time and power.

\bibliography{anthology,custom}
\bibliographystyle{acl_natbib}

\appendix

\clearpage
\onecolumn
\section{Datasets}
\label{sec:Datasets}

In this paper, we have used 6 datasets, including:
GSM8K \citep{cobbeTrainingVerifiersSolve2021}, 
MATH \citep{hendrycksMeasuringMathematicalProblem2021}, 
GSM-HARD \citep{gaoPALProgramaidedLanguage2023}, 
SVAMP \citep{patelAreNLPModels2021}, 
ASDiv \citep{miaoDiverseCorpusEvaluating2020} and 
TAL-SCQ5K.

\noindent
In terms of difficulty, by rough estimation:
$$\text{SVAMP} \approx \text{ASDiV} < \text{GSM8K} \approx \text{GSM-HARD} < \text{TAL-SCQ5k} < \text{MATH}$$
\noindent
with ASDiV as a diversed dataset covering problem types taught in elementary school; SVAMP as a structural modified version of a subset of ASDiv; GSM8K being an immense dataset covering grade school problems, with 2-8 steps; GSM-HARD built upon GSM8K, replacing numbers with less-common large numbers; TAL-SCQ5K containing primary, junior high and high school level mathematical topics; MATH full of challenging competition mathematics problems which requires a strong mathematical background to perform well on. Among which, MATH dataset and TAL-SCQ5K dataset further process notations of difficulty levels.

\begin{xltabular}{\textwidth}{|m{2.5cm}<{\centering}|m{2.5cm}<{\centering}|X|}
\hline
\textbf{Dataset} & \textbf{Num} & \textbf{ Example Q\&A}\\
\hline
\endfirsthead
\hline
\textbf{Dataset} & \textbf{Num} & \textbf{ Example Q\&A}\\
\hline
\endhead
\multirow{11}{2.5cm}{\centering GSM8K} & 
\multirow{11}{2.5cm}{\centering Train: 7473 \\ Test: 1319} & 
\textbf{question:} 
In a dance class of 20 students, 20\% enrolled in contemporary dance, 25\% of the remaining enrolled in jazz dance, and the rest enrolled in hip-hop dance. What percentage of the entire students enrolled in hip-hop dance?  \newline
\textbf{answer:} 
There are 20 x 20/100 = <<20*20/100=4>>4 students who enrolled in contemporary dance. So, 20 - 4 = <<20-4=16>>16 students are enrolled in either jazz or hip-hop dance. There are 16 x 25/100 = <<16*25/100=4>>4 students who enrolled in jazz dance. Hence, 16 - 4 = <<16-4=12>>12 students enrolled in hip-hop dance. This is 12/20 x 100\% = 60\% of the entire students. \#\#\#\# 60 
\\
\hline
\multirow{7}{2.5cm}{\centering MATH} & 
\multirow{7}{2.5cm}{\centering Train: 7500 \\ Test: 5000} & 
\textbf{question:} 
How many vertical asymptotes does the graph of \$ y=\textbackslash frac \{2\}\{x\^{}2+x-6\}\$ have? \newline
\textbf{answer:} 
The denominator of the rational function factors into \$x\^{}2+x-6=(x-2)(x+3)\$. Since the numerator is always nonzero, there is a vertical asymptote whenever the denominator is \$0\$, which occurs for \$x = 2\$ and \$x = -3\$.  Therefore, the graph has \$\textbackslash boxed\{2\}\$ vertical asymptotes.
\\
\hline
\multirow{12}{2.5cm}{\centering GSM-HARD} & 
\multirow{12}{*}{Test: 1319} & 
\textbf{input:} 
A robe takes 2287720 bolts of blue fiber and half that much white fiber.  How many bolts in total does it take? \newline
\textbf{code:} \newline
def solution():\newline   \hspace*{0.5cm}    """A robe takes 2 bolts of blue fiber and half that much white fiber.  How many bolts in total does it take?"""\newline   \hspace*{0.5cm}    blue\_fiber = 2287720\newline   \hspace*{0.5cm}    white\_fiber = blue\_fiber / 2\newline   \hspace*{0.5cm}    total\_fiber = blue\_fiber + white\_fiber\newline   \hspace*{0.5cm}    result = total\_fiber\newline   \hspace*{0.5cm}    return result \newline
\textbf{target:} 
3431580.0
\\
\hline
\multirow{8}{2.5cm}{\centering SVAMP} & 
\multirow{8}{*}{Test: 1000} & 
\textbf{Body:} 
The Razorback t-shirt shop makes \$ 78 dollars off each t-shirt sold. During the Arkansas game and the Texas tech game they sold a total of 186 t-shirts. If they sold 172 t-shirts during the Arkansas game \newline
\textbf{Question:} 
How much money did they make from selling the t-shirts during the Texas tech game? \newline
\textbf{Equation:} 
( 78.0 * ( 186.0 - 172.0 ) ) \newline
\textbf{Answer:} 
1092.0
\\
\hline
\multirow{8}{2.5cm}{\centering ASDiv} & 
\multirow{8}{*}{Test: 2215} & 
\textbf{body:} 
Robert wants to practice goal kicks for soccer. He decides to have 98 kicks before going home from the park. He takes 43 kicks before taking a break to get a drink of water. He then takes another 36 kicks. \newline
\textbf{question:} 
How many more kicks does he need to make before he goes home? \newline
\textbf{equation:} 
98-43-36=19 \newline
\textbf{answer:} 
19 (kicks)
\\
\hline
\multirow{6}{2.5cm}{\centering TAL-SCQ} & 
\multirow{6}{*}{5000} & 
\textbf{problem:} 
If \$n\$ is an even positive integer, the double factorial notation \$n!!\$ represents the product of all the even integers from \$2\$ to \$n\$. For example, \$8!!=2\textbackslash\textbackslash cdot4\textbackslash\textbackslash cdot6\textbackslash\textbackslash cdot8\$. What is the units digit of the following sum?  \$2!!+4!!+6!!+\textbackslash\textbackslash cdot\textbackslash\textbackslash cdot\textbackslash\textbackslash cdot+2018!!+2020!!+2022!!\$ \newline
\textbf{solution:} 
Answer: \$\$2\$\$ 
\\
\hline
\caption{
Examples of datasets in their original format.
}
\end{xltabular}
\twocolumn

\clearpage
\onecolumn

\section{Relationships of k \& N}
\label{appendix:relation_of_k&N}

\begin{figure*}[ht]
  \centering
  \includegraphics[width=1\textwidth]{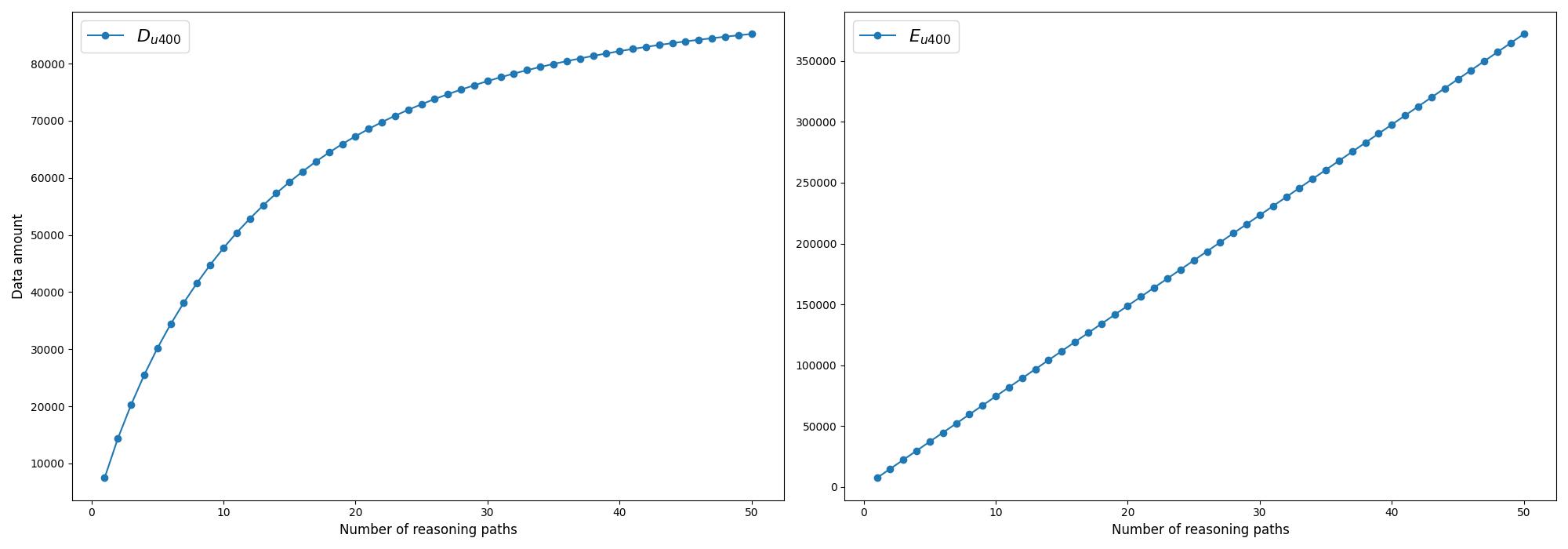}
  \caption{The relationships of k \& N.}
  \label{fig:knGraph1}
\end{figure*}

Figure \ref{fig:knGraph1} illustrates the relationships of the number of reasoning paths and the data amounts of the respective $D_{u400}$ and $E_{u400}$.

We select multiple points from $D_{u400}$ at regular intervals based on data amount. Simultaneously, we choose corresponding points from $E_{u400}$ with similar data amounts to ensure consistence. The statistic that the relationships of the number of reasoning paths and the data amount is detailed in Table \ref{tab:amount1} and \ref{tab:amount2}.

\begin{table*}[h]
\centering{%
\small
\begin{tabular}{c|cccccccccccc}
\hline
$k$ & 1 & 2 & 3 & 5 & 7 & 9 & 12 & 15 & 20 & 27 & 40 & $\infty$  \\
\hline
$N$ & 7457 & 14344 & 20225 & 30179 & 38150 & 44771 & 52857 & 59261 & 67281 & 74643 & 82180 & 89530 \\
\hline
\end{tabular}
}
\caption{Extract subsets from relationships of k \& N of $D_{u400}$ for experiments.}
\label{tab:amount1}
\end{table*}

\begin{table*}[h]
\centering{%
\small
\begin{tabular}{c|cccccccc}
\hline
$k$ & 1 & 2 & 4 & 8 & 12 & 24 & 36 & 48  \\
\hline
$N$ & 7457 & 14911 & 29810 & 59603 & 89386 & 178707 & 268003 & 357295\\
\hline
\end{tabular}
}
\caption{Extract subsets from relationships of k \& N of $E_{u400}$ for experiments.}
\label{tab:amount2}
\end{table*}

\newpage
\section{Detailed Experiment Setting}
\label{subsec: Experiment Setup}

\textbf{Generate Deduplicated Datasets} 

We spent 4 days generating both $D_{u400}$, $D_M$ and the deduplicated dataset of TAL-SCQ in Section \ref{subsec:3.3}, \ref{subsec:4.2} and \ref{subsec:4.5} which is formed by employing four pre-trained models: ToRA-CODE 7B/13B/34B and ToRA 70B on the GSM8K, MATH and TAL-SCQ seperately, these models sample 100 reasoning paths each with temperature 0.9.  \\ 
\textbf{Training Models}

We conducted SFT on Code LLaMA 7B using various deduplicated dataset and their subsets in Section \ref{subsec:3.4}, \ref{subsec:3.5}, \ref{subsec:4.3} and \ref{subsec:4.4}. Addtionally we conducted SFT on LLaMA-2 7B/13B for a horizontal comparison in Section \ref{subsec:4.5}. 

We used a learning rate of 2e-5 with a $3\%$ warm-up period for 1 epoch and a global batch size of 128 on NVIDIA A100 40G GPUs. We trained all models with DeepSpeed ZeRO Stage3 and Flash-Attention 2. 

Apart from validating the effectiveness of the deduplication algorithm, where the random selection process with seeds set to 0 and 42 and then averaging the inference results, all other training and inference processes used a seed of 0.

The training sessions were completed within 1 day, with an average training duration of approximately 5 hours. The average evaluation time is less than 10 minutes. 


\clearpage

\section{Case Study: Actual Distinct Solutions}
\label{appendix:minimal_sets}


To validate the effectiveness of deduplication and using clustering as a filter, we conduct a case study focusing on the relationship of reasoning paths and their problems' actual distinct solutions.

In the deduplicated subset $D_{u400}$ of the GSM8K dataset, we select the first question that has more than 15 reasoning paths, which has 43 reasoning paths for this problem in fact.
Next, we utilize random selection and clustering as a filter to derive the subsets $D^{k=15}_{u400}$ and $D^{cluster,k=15}_{u400}$.
We then separately analyze the 15 reasoning paths in these two subsets for the corresponding problem to categorize their actual distinct solutions on Table \ref{tab:clusters} and \ref{tab:random select}.



The question is formulated as follows:

\definecolor{grayishred}{rgb}{0.7, 0.3, 0.3}
\textit{\textcolor{grayishred}{Tina makes \$18.00 an hour.  If she works more than 8 hours per shift, she is eligible for overtime, which is paid by your hourly wage + 1/2 your hourly wage.  If she works 10 hours every day for 5 days, how much money does she make?}} 

Upon human analysis of this question, 10 distinct solutions have been summarized. These solutions are categorized based on whether the default daily salary is the same, whether the default working hours exceed the regular working hours, and how the total salary is calculated. 
\begin{enumerate}
  \item[A] Assuming that the daily wage is different, calculate the total wage by iterating over each day's wage and summing them in a loop. 
  \item[B] Assuming that the daily wage remains the same and that the working hours exceed the regular hours. Calculate the regular wage and the overtime wage for five days separately, then sum them up. 
  \item[C] Assuming that the daily wage remains the same and that the working hours exceed the regular hours. Calculate the regular hours worked and the overtime hours worked for five days separately, then sum the total wages. 
  \item[D] Assuming that the daily wage remains the same and that the working hours exceed the regular hours. Calculate daily wages, then sum them up.
  \item[E] Assuming that the daily wage remains the same and that the Min and Max functions are used to avoid situations where the working hours are smaller than the regular hours. Calculate the regular wage and the overtime wage for five days separately, then sum them up. 
  \item[F] Assuming that the daily wage remains the same and that the Min and Max functions are used to avoid situations where the working hours are smaller than the regular hours. Calculate the regular hours worked and the overtime hours worked for five days separately, then sum the total wages.
  \item[G] Assuming that the daily wage remains the same and that the Min and Max functions are used to avoid situations where the working hours are smaller than the regular hours. Calculate daily wages, then sum them up.
  \item[H] Assuming that the daily wage remains the same and determine if the working hours exceed the regular hours. Calculate the regular wage and the overtime wage for five days separately, then sum them up. 
  \item[I] Assuming that the daily wage remains the same and determine if the working hours exceed the regular hours. Calculate the regular hours worked and the overtime hours worked for five days separately, then sum the total wages.
  \item[J]  Assuming that the daily wage remains the same and determine if the working hours exceed the regular hours. Calculate daily wages, then sum them up.
\end{enumerate}

\onecolumn
\begin{longtable}{c p{0.75\textwidth}c} \caption{Examples for Clustering as a Filter's Actual Solutions.} \\
\hline
\textbf{ID} & \textbf{Completion} & \textbf{Type} \\
\hline
\endfirsthead


& \texttt{\textasciigrave\textasciigrave\textasciigrave}python\\&
def total\_salary():\\&   
\hspace*{0.5cm} \textcolor{grayishred}{"""Tina makes \$18.00 an hour. If she works more than 8 hours per shift, she is eligible for overtime, which is paid by your hourly wage + 1/2 your hourly wage. If she works 10 hours every day for 5 days, how much money does she make?"""}\\&   
\hspace*{0.5cm} hourly\_wage = 18.00\\&   
\hspace*{0.5cm} hours\_per\_shift = 10\\&   
\hspace*{0.5cm} num\_days = 5\\&   
\hspace*{0.5cm} overtime\_hours = hours\_per\_shift - 8\\1&   \hspace*{0.5cm} overtime\_wage = hourly\_wage + 0.5 * hourly\_wage&J\\&   
\hspace*{0.5cm} salary\_per\_day = hourly\_wage * 8 + overtime\_wage * overtime\_hours\\&   
\hspace*{0.5cm} total\_salary = salary\_per\_day * num\_days\\& \hspace*{0.5cm} return total\_salary\\& total\_salary = total\_salary()\\& print(total\_salary)\\& 	\texttt{\textasciigrave\textasciigrave\textasciigrave} \\& \texttt{\textasciigrave\textasciigrave\textasciigrave} output\\& 990.0\\& \texttt{\textasciigrave\textasciigrave\textasciigrave} \\& Tina makes \fcolorbox{red}{white}{$990.0$} dollars. 
&\\
\hline

& \texttt{\textasciigrave\textasciigrave\textasciigrave}python\\&
def total\_pay():\\&
\hspace*{0.5cm} \textcolor{grayishred}{"""Tina makes \$18.00 an hour. If she works more than 8 hours per shift, she is eligible for overtime, which is paid by your hourly wage + 1/2 your hourly wage. If she works 10 hours every day for 5 days, how much money does she make?"""}\\&
\hspace*{0.5cm}hourly\_wage = 18.00\\&
\hspace*{0.5cm}hours\_per\_shift = 10\\&
\hspace*{0.5cm}num\_shifts = 5\\&
\hspace*{0.5cm}overtime\_hours = hours\_per\_shift - 8\\&
\hspace*{0.5cm}overtime\_wage = hourly\_wage + 0.5 * hourly\_wage\\2&
\hspace*{0.5cm}pay\_for\_overtime = overtime\_hours * overtime\_wage * num\_shifts&B\\&
\hspace*{0.5cm}pay\_for\_regular\_hours = (8 * hourly\_wage) * num\_shifts\\&
\hspace*{0.5cm}total\_pay = pay\_for\_overtime + pay\_for\_regular\_hours\\&
\hspace*{0.5cm}return total\_pay\\&
total\_pay = total\_pay()\\&
print(total\_pay)\\&
\texttt{\textasciigrave\textasciigrave\textasciigrave}\\&
\texttt{\textasciigrave\textasciigrave\textasciigrave} output\\&
990.0\\&
\texttt{\textasciigrave\textasciigrave\textasciigrave}\\&
Tina makes \fcolorbox{red}{white}{$990.0$} dollars.
&\\
\hline
& \texttt{\textasciigrave\textasciigrave\textasciigrave}python\\&
def calculate\_earnings():\\&
\hspace*{0.5cm}\textcolor{grayishred}{"""Tina makes \$18.00 an hour. If she works more than 8 hours per shift, she is eligible for overtime, which is paid by your hourly wage + 1/2 your hourly wage. If she works 10 hours every day for 5 days, how much money does she make?"""}\\&
\hspace*{0.5cm}hourly\_wage = 18.00\\&
\hspace*{0.5cm}overtime\_hours = 10 - 8  \# hours worked in excess of 8 hours\\&
\hspace*{0.5cm}overtime\_wage = hourly\_wage + 0.5 * hourly\_wage  \# overtime wage = hourly wage + half hourly wage\\3&
\hspace*{0.5cm}daily\_pay = 8 * hourly\_wage + overtime\_hours * overtime\_wage&D\\&
\hspace*{0.5cm}total\_pay = daily\_pay * 5  \# total pay for 5 days\\&
\hspace*{0.5cm}return total\_pay\\&
total\_pay = calculate\_earnings()\\&
print(total\_pay)\\&
\texttt{\textasciigrave\textasciigrave\textasciigrave}\\&
\texttt{\textasciigrave\textasciigrave\textasciigrave} output\\&
990.0\\&
\texttt{\textasciigrave\textasciigrave\textasciigrave}\\&
Tina makes \fcolorbox{red}{white}{$990.0$} dollars.\\
\hline
& \texttt{\textasciigrave\textasciigrave\textasciigrave}python\\&
def total\_pay():\\&
\hspace*{0.5cm}\textcolor{grayishred}{"""Tina makes \$18.00 an hour. If she works more than 8 hours per shift, she is eligible for overtime, which is paid by your hourly wage + 1/2 your hourly wage. If she works 10 hours every day for 5 days, how much money does she make?"""}\\&
\hspace*{0.5cm}hourly\_wage = 18.00\\&
\hspace*{0.5cm}hours\_per\_shift = 10\\&
\hspace*{0.5cm}overtime\_threshold = 8\\&
\hspace*{0.5cm}num\_shifts = 5\\&
\hspace*{0.5cm}total\_pay = 0\\&
\hspace*{0.5cm}for i in range(num\_shifts):\\&
\hspace*{1cm}if hours\_per\_shift <= overtime\_threshold:\\4&
\hspace*{1cm}\hspace*{0.5cm}total\_pay += hourly\_wage * hours\_per\_shift&A\\&
\hspace*{1cm}else:\\&
\hspace*{1cm}\hspace*{0.5cm}overtime\_hours = hours\_per\_shift - overtime\_threshold\\&
\hspace*{1cm}\hspace*{0.5cm}regular\_hours = overtime\_threshold\\&
\hspace*{1cm}\hspace*{0.5cm}total\_pay += (hourly\_wage * regular\_hours) + (hourly\_wage * overtime\_hours * 1.5)\\&
\hspace*{0.5cm}return total\_pay\\&
total\_pay = total\_pay()\\&
print(total\_pay)\\&
\texttt{\textasciigrave\textasciigrave\textasciigrave}\\&
\texttt{\textasciigrave\textasciigrave\textasciigrave} output\\&
990.0\\&
\texttt{\textasciigrave\textasciigrave\textasciigrave}\\&
Tina makes \fcolorbox{red}{white}{$990.0$} dollars.
&\\
\hline
& \texttt{\textasciigrave\textasciigrave\textasciigrave}python\\&
def calculate\_pay():\\&
\hspace*{0.5cm}\textcolor{grayishred}{"""Tina makes \$18.00 an hour. If she works more than 8 hours per shift, she is eligible for overtime, which is paid by your hourly wage + 1/2 your hourly wage. If she works 10 hours every day for 5 days, how much money does she make?"""}\\&
\hspace*{0.5cm}hourly\_wage = 18.00\\&
\hspace*{0.5cm}hours\_per\_shift = 10\\&
\hspace*{0.5cm}num\_days = 5\\&
\hspace*{0.5cm}overtime\_hours = hours\_per\_shift - 8\\&
\hspace*{0.5cm}overtime\_pay = hourly\_wage + 0.5 * hourly\_wage\\5&
\hspace*{0.5cm}regular\_pay = hourly\_wage * 8&D\\&
\hspace*{0.5cm}total\_pay = (overtime\_hours * overtime\_pay + regular\_pay) * num\_days\\&
\hspace*{0.5cm}return total\_pay\\&
total\_pay = calculate\_pay()\\&
print(total\_pay)\\&
\texttt{\textasciigrave\textasciigrave\textasciigrave}\\&
\texttt{\textasciigrave\textasciigrave\textasciigrave} output\\&
990.0\\&
\texttt{\textasciigrave\textasciigrave\textasciigrave}\\&
Tina makes \fcolorbox{red}{white}{$990.0$} dollars.
&\\
\hline
& \texttt{\textasciigrave\textasciigrave\textasciigrave}python\\&
def total\_earnings():\\&
\hspace*{0.5cm}\textcolor{grayishred}{"""Tina makes \$18.00 an hour. If she works more than 8 hours per shift, she is eligible for overtime, which is paid by your hourly wage + 1/2 your hourly wage. If she works 10 hours every day for 5 days, how much money does she make?"""}\\&
\hspace*{0.5cm}hourly\_wage = 18.00\\&
\hspace*{0.5cm}hours\_per\_shift = 10\\&
\hspace*{0.5cm}num\_days = 5\\&
\hspace*{0.5cm}overtime\_hours = hours\_per\_shift - 8 if hours\_per\_shift > 8 else 0\\6&
\hspace*{0.5cm}earnings = (hourly\_wage * 8) + (hourly\_wage * overtime\_hours * 1.5)&J\\&
\hspace*{0.5cm}total\_earnings = earnings * num\_days\\&
\hspace*{0.5cm}return total\_earnings\\&
total\_earnings = total\_earnings()\\&
print(total\_earnings)\\&
\texttt{\textasciigrave\textasciigrave\textasciigrave}\\&
\texttt{\textasciigrave\textasciigrave\textasciigrave} output\\&
990.0\\&
\texttt{\textasciigrave\textasciigrave\textasciigrave}\\&
Tina makes \fcolorbox{red}{white}{$990.0$} dollars.
&\\
\hline
& \texttt{\textasciigrave\textasciigrave\textasciigrave}python\\& total\_earnings():\\&
\hspace*{0.5cm}\textcolor{grayishred}{"""Tina makes \$18.00 an hour. If she works more than 8 hours per shift, she is eligible for overtime, which is paid by your hourly wage + 1/2 your hourly wage. If she works 10 hours every day for 5 days, how much money does she make?"""}\\&
\hspace*{0.5cm}hourly\_wage = 18.00\\&
\hspace*{0.5cm}overtime\_wage = hourly\_wage + 0.5 * hourly\_wage\\&
\hspace*{0.5cm}hours\_per\_shift = 10\\&
\hspace*{0.5cm}num\_days = 5\\7&
\hspace*{0.5cm}total\_hours = hours\_per\_shift * num\_days&C\\&
\hspace*{0.5cm}overtime\_hours = total\_hours - 8 * num\_days\\&
\hspace*{0.5cm}earnings = (8 * num\_days * hourly\_wage) + (overtime\_hours * overtime\_wage)\\&
\hspace*{0.5cm}return earnings\\&
total\_earnings = total\_earnings()\\&
print(total\_earnings)\\&
\texttt{\textasciigrave\textasciigrave\textasciigrave}\\&
\texttt{\textasciigrave\textasciigrave\textasciigrave} output\\&
990.0\\&
\texttt{\textasciigrave\textasciigrave\textasciigrave}\\&
Tina makes \fcolorbox{red}{white}{$990.0$} dollars.
&\\
\hline
& \texttt{\textasciigrave\textasciigrave\textasciigrave}python\\& def total\_pay():\\&
\hspace*{0.5cm}\textcolor{grayishred}{"""Tina makes \$18.00 an hour. If she works more than 8 hours per shift, she is eligible for overtime, which is paid by your hourly wage + 1/2 your hourly wage. If she works 10 hours every day for 5 days, how much money does she make?"""}\\&
\hspace*{0.5cm}hourly\_wage = 18.00\\&
\hspace*{0.5cm}hours\_per\_day = 10\\&
\hspace*{0.5cm}num\_days = 5\\&
\hspace*{0.5cm}overtime\_hours = hours\_per\_day - 8 if hours\_per\_day > 8 else 0\\8&
\hspace*{0.5cm}overtime\_pay = overtime\_hours * (hourly\_wage + 0.5 * hourly\_wage)&J\\&
\hspace*{0.5cm}regular\_pay = (8 * hourly\_wage) + overtime\_pay\\&
\hspace*{0.5cm}total\_pay = regular\_pay * num\_days\\&
\hspace*{0.5cm}return total\_pay\\&

total\_pay = total\_pay()\\&
print(total\_pay)\\&
\texttt{\textasciigrave\textasciigrave\textasciigrave}\\&
\texttt{\textasciigrave\textasciigrave\textasciigrave} output\\&
990.0\\&
\texttt{\textasciigrave\textasciigrave\textasciigrave}\\&
Tina makes \fcolorbox{red}{white}{$990.0$} dollars.
&\\
\hline
& \texttt{\textasciigrave\textasciigrave\textasciigrave}python\\& def total\_earnings():\\&
\hspace*{0.5cm}\textcolor{grayishred}{"""Tina makes \$18.00 an hour. If she works more than 8 hours per shift, she is eligible for overtime, which is paid by your hourly wage + 1/2 your hourly wage. If she works 10 hours every day for 5 days, how much money does she make?"""}\\&
\hspace*{0.5cm}hourly\_wage = 18\\&
\hspace*{0.5cm}hours\_per\_shift = 10\\&
\hspace*{0.5cm}num\_days = 5\\&
\hspace*{0.5cm}overtime\_hours = hours\_per\_shift - 8\\9&
\hspace*{0.5cm}overtime\_wage = hourly\_wage + 0.5 * hourly\_wage&D\\&
\hspace*{0.5cm}earnings = (8 * hourly\_wage + overtime\_hours * overtime\_wage) * num\_days\\&
\hspace*{0.5cm}return earnings\\&
total\_earnings = total\_earnings()\\&
print(total\_earnings)\\&
\texttt{\textasciigrave\textasciigrave\textasciigrave}\\&
\texttt{\textasciigrave\textasciigrave\textasciigrave} output\\&
990.0\\&
\texttt{\textasciigrave\textasciigrave\textasciigrave}\\&
Tina makes \fcolorbox{red}{white}{$990.0$} dollars.
&\\
\hline
& \texttt{\textasciigrave\textasciigrave\textasciigrave}python\\&
def calculate\_pay():\\&
\hspace*{0.5cm}\textcolor{grayishred}{"""Tina makes \$18.00 an hour. If she works more than 8 hours per shift, she is eligible for overtime, which is paid by your hourly wage + 1/2 your hourly wage. If she works 10 hours every day for 5 days, how much money does she make?"""}\\&
\hspace*{0.5cm}hourly\_wage = 18\\&
\hspace*{0.5cm}hours\_per\_shift = 10\\&
\hspace*{0.5cm}days\_worked = 5\\&
\hspace*{0.5cm}overtime\_hours = hours\_per\_shift - 8\\&
\hspace*{0.5cm}overtime\_wage = hourly\_wage + 0.5 * hourly\_wage\\&
\hspace*{0.5cm}regular\_pay = hourly\_wage * 8\\10&
\hspace*{0.5cm}overtime\_pay = overtime\_wage * overtime\_hours&D\\&
\hspace*{0.5cm}total\_pay = (regular\_pay + overtime\_pay) * days\_worked\\&
\hspace*{0.5cm}return total\_pay\\&
total\_pay = calculate\_pay()\\&
print(total\_pay)\\&
\texttt{\textasciigrave\textasciigrave\textasciigrave}\\&
\texttt{\textasciigrave\textasciigrave\textasciigrave} output\\&
990.0\\&
\texttt{\textasciigrave\textasciigrave\textasciigrave}\\&
Tina makes \fcolorbox{red}{white}{$990.0$} dollars.
&\\
\hline
& \texttt{\textasciigrave\textasciigrave\textasciigrave}python\\&
def total\_earnings():\\&
\hspace*{0.5cm}\textcolor{grayishred}{"""Tina makes \$18.00 an hour. If she works more than 8 hours per shift, she is eligible for overtime, which is paid by your hourly wage + 1/2 your hourly wage. If she works 10 hours every day for 5 days, how much money does she make?"""}\\&
\hspace*{0.5cm}hourly\_wage = 18.00\\&
\hspace*{0.5cm}overtime\_wage = hourly\_wage + 0.5 * hourly\_wage\\&
\hspace*{0.5cm}hours\_per\_shift = 10\\&
\hspace*{0.5cm}num\_shifts = 5\\11&
\hspace*{0.5cm}earnings\_straight\_time = hourly\_wage * 8 * num\_shifts&B\\&
\hspace*{0.5cm}earnings\_overtime = overtime\_wage * (hours\_per\_shift - 8) * num\_shifts\\&
\hspace*{0.5cm}total\_earnings = earnings\_straight\_time + earnings\_overtime\\&
\hspace*{0.5cm}return total\_earnings\\&
total\_earnings = total\_earnings()\\&
print(total\_earnings)\\&
\texttt{\textasciigrave\textasciigrave\textasciigrave}\\&
\texttt{\textasciigrave\textasciigrave\textasciigrave} output\\&
990.0\\&
\texttt{\textasciigrave\textasciigrave\textasciigrave}\\&
Tina makes \fcolorbox{red}{white}{$990.0$} dollars.
&\\
\hline
& \texttt{\textasciigrave\textasciigrave\textasciigrave}python\\&
def calculate\_salary():\\&
\hspace*{0.5cm}\textcolor{grayishred}{"""Tina makes \$18.00 an hour. If she works more than 8 hours per shift, she is eligible for overtime, which is paid by your hourly wage + 1/2 your hourly wage. If she works 10 hours every day for 5 days, how much money does she make?"""}\\&
\hspace*{0.5cm}hourly\_wage = 18.00\\&
\hspace*{0.5cm}overtime\_wage = hourly\_wage + 0.5 * hourly\_wage\\&
\hspace*{0.5cm}hours\_per\_shift = 10\\&
\hspace*{0.5cm}days = 5\\&
\hspace*{0.5cm}regular\_hours = min(hours\_per\_shift, 8)\\12&
\hspace*{0.5cm}overtime\_hours = max(hours\_per\_shift - 8, 0)&E\\&
\hspace*{0.5cm}regular\_pay = regular\_hours * hourly\_wage * days\\&
\hspace*{0.5cm}overtime\_pay = overtime\_hours * overtime\_wage * days\\&
\hspace*{0.5cm}total\_pay = regular\_pay + overtime\_pay\\&
\hspace*{0.5cm}return total\_pay\\&
total\_pay = calculate\_salary()\\&
print(total\_pay)\\&
\texttt{\textasciigrave\textasciigrave\textasciigrave}\\&
\texttt{\textasciigrave\textasciigrave\textasciigrave} output\\&
990.0\\&
\texttt{\textasciigrave\textasciigrave\textasciigrave}\\&
Tina makes \fcolorbox{red}{white}{$990.0$} dollars.
&\\
\hline
& \texttt{\textasciigrave\textasciigrave\textasciigrave}python\\&
def total\_pay():\\&
\hspace*{0.5cm}\textcolor{grayishred}{"""Tina makes \$18.00 an hour. If she works more than 8 hours per shift, she is eligible for overtime, which is paid by your hourly wage + 1/2 your hourly wage. If she works 10 hours every day for 5 days, how much money does she make?"""}\\&
\hspace*{0.5cm}hourly\_wage = 18.00\\&
\hspace*{0.5cm}overtime\_wage = hourly\_wage + 0.5 * hourly\_wage\\&
\hspace*{0.5cm}hours\_per\_day = 10\\&
\hspace*{0.5cm}num\_days = 5\\&
\hspace*{0.5cm}overtime\_hours = max(hours\_per\_day - 8, 0)\\13&
\hspace*{0.5cm}regular\_hours = min(hours\_per\_day, 8)\\&
\hspace*{0.5cm}total\_pay = (regular\_hours * hourly\_wage + overtime\_hours * overtime\_wage) * num\_days&G\\&
\hspace*{0.5cm}return total\_pay\\&
total\_pay = total\_pay()\\&
print(total\_pay)\\&
\texttt{\textasciigrave\textasciigrave\textasciigrave}\\&
\texttt{\textasciigrave\textasciigrave\textasciigrave} output\\&
990.0\\&
\texttt{\textasciigrave\textasciigrave\textasciigrave}\\&
Tina makes \fcolorbox{red}{white}{$990.0$} dollars.
&\\
\hline
& \texttt{\textasciigrave\textasciigrave\textasciigrave}python\\&
def total\_earnings():\\&
\hspace*{0.5cm}\textcolor{grayishred}{"""Tina makes \$18.00 an hour. If she works more than 8 hours per shift, she is eligible for overtime, which is paid by your hourly wage + 1/2 your hourly wage. If she works 10 hours every day for 5 days, how much money does she make?"""}\\&
\hspace*{0.5cm}hourly\_wage = 18.00\\&
\hspace*{0.5cm}overtime\_hours = 10\\&
\hspace*{0.5cm}days = 5\\&
\hspace*{0.5cm}regular\_hours = 8\\&
\hspace*{0.5cm}regular\_earnings = hourly\_wage * regular\_hours * days\\14&
\hspace*{0.5cm}overtime\_earnings = (hourly\_wage + 0.5 * hourly\_wage) * (overtime\_hours - regular\_hours) * days&B\\&
\hspace*{0.5cm}total\_earnings = regular\_earnings + overtime\_earnings\\&
\hspace*{0.5cm}return total\_earnings\\&
total\_earnings = total\_earnings()\\&
print(total\_earnings)\\&
\texttt{\textasciigrave\textasciigrave\textasciigrave}\\&
\texttt{\textasciigrave\textasciigrave\textasciigrave} output\\&
990.0\\&
\texttt{\textasciigrave\textasciigrave\textasciigrave}\\&
Tina makes \fcolorbox{red}{white}{$990.0$} dollars.
&\\
\hline
& \texttt{\textasciigrave\textasciigrave\textasciigrave}python\\&
def calculate\_pay():\\&
\hspace*{0.5cm}\textcolor{grayishred}{"""Tina makes \$18.00 an hour. If she works more than 8 hours per shift, she is eligible for overtime, which is paid by your hourly wage + 1/2 your hourly wage. If she works 10 hours every day for 5 days, how much money does she make?"""}\\&
\hspace*{0.5cm}hourly\_wage = 18.00\\&
\hspace*{0.5cm}hours\_per\_shift = 10\\&
\hspace*{0.5cm}num\_shifts = 5\\&
\hspace*{0.5cm}overtime\_hours = hours\_per\_shift - 8\\15&
\hspace*{0.5cm}overtime\_wage = hourly\_wage + 0.5 * hourly\_wage&D\\&
\hspace*{0.5cm}pay = (8 * hourly\_wage + overtime\_hours * overtime\_wage) * num\_shifts\\&
\hspace*{0.5cm}return pay\\&
total\_pay = calculate\_pay()\\&
print(total\_pay)\\&
\texttt{\textasciigrave\textasciigrave\textasciigrave}\\&
\texttt{\textasciigrave\textasciigrave\textasciigrave} output\\&
990.0\\&
\texttt{\textasciigrave\textasciigrave\textasciigrave}\\&
Tina makes \fcolorbox{red}{white}{$990.0$} dollars.
&\\
\hline
\label{tab:clusters}
\end{longtable}

\begin{longtable}{c p{0.75\textwidth}c} 
\caption{Examples for Random Selection's Actual Solutions.} \\
\hline
\textbf{ID} & \textbf{Completion} & \textbf{Type} \\
\hline
\endfirsthead
& \texttt{\textasciigrave\textasciigrave\textasciigrave}python\\&
def total\_salary():\\&
\hspace*{0.5cm}\textcolor{grayishred}{"""Tina makes \$18.00 an hour. If she works more than 8 hours per shift, she is eligible for overtime, which is paid by your hourly wage + 1/2 your hourly wage. If she works 10 hours every day for 5 days, how much money does she make?"""}\\&
\hspace*{0.5cm}hourly\_wage = 18.00\\&
\hspace*{0.5cm}hours\_per\_shift = 10\\&
\hspace*{0.5cm}num\_days = 5\\&
\hspace*{0.5cm}overtime\_hours = hours\_per\_shift - 8\\1&
\hspace*{0.5cm}overtime\_wage = hourly\_wage + 0.5 * hourly\_wage&D\\&
\hspace*{0.5cm}salary\_per\_day = hourly\_wage * 8 + overtime\_wage * overtime\_hours\\&
\hspace*{0.5cm}total\_salary = salary\_per\_day * num\_days\\&
\hspace*{0.5cm}return total\_salary\\&
total\_salary = total\_salary()\\&
print(total\_salary)\\&
\texttt{\textasciigrave\textasciigrave\textasciigrave}\\&
\texttt{\textasciigrave\textasciigrave\textasciigrave} output\\&
990.0\\&
\texttt{\textasciigrave\textasciigrave\textasciigrave}\\&
Tina makes \fcolorbox{red}{white}{$990.0$} dollars.
&\\
\hline
& \texttt{\textasciigrave\textasciigrave\textasciigrave}python\\&
def total\_pay():\\&
\hspace*{0.5cm}\textcolor{grayishred}{"""Tina makes \$18.00 an hour. If she works more than 8 hours per shift, she is eligible for overtime, which is paid by your hourly wage + 1/2 your hourly wage. If she works 10 hours every day for 5 days, how much money does she make?"""}\\&
\hspace*{0.5cm}hourly\_wage = 18.00\\&
\hspace*{0.5cm}hours\_per\_shift = 10\\&
\hspace*{0.5cm}overtime\_threshold = 8\\&
\hspace*{0.5cm}num\_shifts = 5\\&
\hspace*{0.5cm}total\_pay = 0\\&
\hspace*{0.5cm}for i in range(num\_shifts):\\&
\hspace*{0.5cm}\hspace*{0.5cm}if hours\_per\_shift <= overtime\_threshold:\\2&
\hspace*{0.5cm}\hspace*{0.5cm}\hspace*{0.5cm}total\_pay += hourly\_wage * hours\_per\_shift&A\\&
\hspace*{0.5cm}\hspace*{0.5cm}else:\\&
\hspace*{0.5cm}\hspace*{0.5cm}\hspace*{0.5cm}overtime\_hours = hours\_per\_shift - overtime\_threshold\\&
\hspace*{0.5cm}\hspace*{0.5cm}\hspace*{0.5cm}regular\_hours = overtime\_threshold\\&
\hspace*{0.5cm}\hspace*{0.5cm}\hspace*{0.5cm}total\_pay += (hourly\_wage * regular\_hours) +\\&
\hspace*{0.5cm}\hspace*{0.5cm}\hspace*{0.5cm}(hourly\_wage * overtime\_hours * 1.5)\\&
\hspace*{0.5cm}return total\_pay\\&
total\_pay = total\_pay()\\&
print(total\_pay)\\&
\texttt{\textasciigrave\textasciigrave\textasciigrave}\\&
\texttt{\textasciigrave\textasciigrave\textasciigrave} output\\&
990.0\\&
\texttt{\textasciigrave\textasciigrave\textasciigrave}\\&
Tina makes \fcolorbox{red}{white}{$990.0$} dollars.
&\\
\hline
& \texttt{\textasciigrave\textasciigrave\textasciigrave}python\\&
def calculate\_pay():\\&
\hspace*{0.5cm}\textcolor{grayishred}{"""Tina makes \$18.00 an hour. If she works more than 8 hours per shift, she is eligible for overtime, which is paid by your hourly wage + 1/2 your hourly wage. If she works 10 hours every day for 5 days, how much money does she make?"""}\\&
\hspace*{0.5cm}hourly\_wage = 18.00\\&
\hspace*{0.5cm}overtime\_wage = 0.5 * hourly\_wage\\&
\hspace*{0.5cm}hours\_per\_shift = 10\\&
\hspace*{0.5cm}num\_shifts = 5\\3&
\hspace*{0.5cm}overtime\_hours = max(hours\_per\_shift - 8, 0)&G\\&
\hspace*{0.5cm}regular\_pay = hourly\_wage * 8\\&
\hspace*{0.5cm}overtime\_pay = (hours\_per\_shift - 8) * (hourly\_wage + overtime\_wage)\\&
\hspace*{0.5cm}total\_pay = (regular\_pay + overtime\_pay) * num\_shifts\\&
\hspace*{0.5cm}return total\_pay\\&
total\_pay = calculate\_pay()\\&
print(total\_pay)\\&
\texttt{\textasciigrave\textasciigrave\textasciigrave}\\&
\texttt{\textasciigrave\textasciigrave\textasciigrave} output\\&
990.0\\&
\texttt{\textasciigrave\textasciigrave\textasciigrave}\\&
Tina makes \fcolorbox{red}{white}{$990.0$} dollars.
&\\
\hline
& \texttt{\textasciigrave\textasciigrave\textasciigrave}python\\&
def calculate\_pay():\\&
\hspace*{0.5cm}\textcolor{grayishred}{"""Tina makes \$18.00 an hour. If she works more than 8 hours per shift, she is eligible for overtime, which is paid by your hourly wage + 1/2 your hourly wage. If she works 10 hours every day for 5 days, how much money does she make?"""}\\&
\hspace*{0.5cm}hourly\_wage = 18.00\\&
\hspace*{0.5cm}hours\_per\_shift = 10\\&
\hspace*{0.5cm}num\_shifts = 5\\&
\hspace*{0.5cm}overtime\_hours = hours\_per\_shift - 8\\4&
\hspace*{0.5cm}overtime\_wage = hourly\_wage + 0.5 * hourly\_wage&D\\&
\hspace*{0.5cm}pay\_for\_shifts = (8 * hourly\_wage) + (overtime\_hours * overtime\_wage)\\&
\hspace*{0.5cm}total\_pay = num\_shifts * pay\_for\_shifts\\&
\hspace*{0.5cm}return total\_pay\\&
total\_pay = calculate\_pay()\\&
print(total\_pay)\\&
\texttt{\textasciigrave\textasciigrave\textasciigrave}\\&
\texttt{\textasciigrave\textasciigrave\textasciigrave} output\\&
990.0\\&
\texttt{\textasciigrave\textasciigrave\textasciigrave}\\&
Tina makes \fcolorbox{red}{white}{$990.0$} dollars.
&\\
\hline
& \texttt{\textasciigrave\textasciigrave\textasciigrave}python\\&
def total\_earnings():\\&
\hspace*{0.5cm}\textcolor{grayishred}{"""Tina makes \$18.00 an hour. If she works more than 8 hours per shift, she is eligible for overtime, which is paid by your hourly wage + 1/2 your hourly wage. If she works 10 hours every day for 5 days, how much money does she make?"""}\\&
\hspace*{0.5cm}hourly\_wage = 18.00\\&
\hspace*{0.5cm}hours\_per\_shift = 10\\&
\hspace*{0.5cm}num\_days = 5\\&
\hspace*{0.5cm}overtime\_hours = hours\_per\_shift - 8\\&
\hspace*{0.5cm}if overtime\_hours > 0:\\&
\hspace*{0.5cm}\hspace*{0.5cm}overtime\_wage = hourly\_wage + 0.5 * hourly\_wage\\5&
\hspace*{0.5cm}\hspace*{0.5cm}overtime\_earnings = overtime\_hours * overtime\_wage * num\_days&H\\&
\hspace*{0.5cm}else:\\&
\hspace*{0.5cm}\hspace*{0.5cm}overtime\_earnings = 0\\&
\hspace*{0.5cm}regular\_hours = 8\\&
\hspace*{0.5cm}regular\_wage = hourly\_wage\\&
\hspace*{0.5cm}regular\_earnings = regular\_hours * regular\_wage * num\_days\\&
\hspace*{0.5cm}total\_earnings = regular\_earnings + overtime\_earnings\\&
\hspace*{0.5cm}return total\_earnings\\&
total\_earnings = total\_earnings()\\&
print(total\_earnings)\\&
\texttt{\textasciigrave\textasciigrave\textasciigrave}\\&
\texttt{\textasciigrave\textasciigrave\textasciigrave} output\\&
990.0\\&
\texttt{\textasciigrave\textasciigrave\textasciigrave}\\&
Tina makes \fcolorbox{red}{white}{$990.0$} dollars.
&\\
\hline
& \texttt{\textasciigrave\textasciigrave\textasciigrave}python\\&
def calculate\_earnings():\\&
\hspace*{0.5cm}\textcolor{grayishred}{"""Tina makes \$18.00 an hour. If she works more than 8 hours per shift, she is eligible for overtime, which is paid by your hourly wage + 1/2 your hourly wage. If she works 10 hours every day for 5 days, how much money does she make?"""}\\&
\hspace*{0.5cm}hourly\_wage = 18.00\\&
\hspace*{0.5cm}overtime\_wage = hourly\_wage + 0.5 * hourly\_wage\\&
\hspace*{0.5cm}hours\_per\_day = 10\\&
\hspace*{0.5cm}days\_per\_week = 5\\&
\hspace*{0.5cm}overtime\_hours = hours\_per\_day - 8  \# overtime is any hours over 8\\6&
\hspace*{0.5cm}base\_earnings = hourly\_wage * 8 * days\_per\_week&B\\&
\hspace*{0.5cm}overtime\_earnings = overtime\_wage * overtime\_hours * days\_per\_week\\&
\hspace*{0.5cm}total\_earnings = base\_earnings + overtime\_earnings\\&
\hspace*{0.5cm}return total\_earnings\\&
total\_earnings = calculate\_earnings()\\&
print(total\_earnings)\\&
\texttt{\textasciigrave\textasciigrave\textasciigrave}\\&
\texttt{\textasciigrave\textasciigrave\textasciigrave} output\\&
990.0\\&
\texttt{\textasciigrave\textasciigrave\textasciigrave}\\&
Tina makes \fcolorbox{red}{white}{$990.0$} dollars.
&\\
\hline
& \texttt{\textasciigrave\textasciigrave\textasciigrave}python\\&
def total\_pay():\\&
\hspace*{0.5cm}\textcolor{grayishred}{"""Tina makes \$18.00 an hour. If she works more than 8 hours per shift, she is eligible for overtime, which is paid by your hourly wage + 1/2 your hourly wage. If she works 10 hours every day for 5 days, how much money does she make?"""}\\&
\hspace*{0.5cm}hourly\_wage = 18.00\\&
\hspace*{0.5cm}hours\_per\_day = 10\\&
\hspace*{0.5cm}num\_days = 5\\&
\hspace*{0.5cm}overtime\_hours = hours\_per\_day - 8 if hours\_per\_day > 8 else 0\\7&
\hspace*{0.5cm}overtime\_pay = overtime\_hours * (hourly\_wage + 0.5 * hourly\_wage)&J\\&
\hspace*{0.5cm}regular\_pay = (8 * hourly\_wage) + overtime\_pay\\&
\hspace*{0.5cm}total\_pay = regular\_pay * num\_days\\&
\hspace*{0.5cm}return total\_pay\\&
total\_pay = total\_pay()\\&
print(total\_pay)\\&
\texttt{\textasciigrave\textasciigrave\textasciigrave}\\&
\texttt{\textasciigrave\textasciigrave\textasciigrave} output\\&
990.0\\&
\texttt{\textasciigrave\textasciigrave\textasciigrave}\\&
Tina makes \fcolorbox{red}{white}{$990.0$} dollars.
&\\
\hline
& \texttt{\textasciigrave\textasciigrave\textasciigrave}python\\&
def calculate\_earnings():\\&
\hspace*{0.5cm}\textcolor{grayishred}{"""Tina makes \$18.00 an hour. If she works more than 8 hours per shift, she is eligible for overtime, which is paid by your hourly wage + 1/2 your hourly wage. If she works 10 hours every day for 5 days, how much money does she make?"""}\\&
\hspace*{0.5cm}hourly\_wage = 18.00\\&
\hspace*{0.5cm}hours\_per\_shift = 10\\&
\hspace*{0.5cm}num\_days = 5\\&
\hspace*{0.5cm}overtime\_hours = hours\_per\_shift - 8\\&
\hspace*{0.5cm}overtime\_wage = hourly\_wage + 0.5 * hourly\_wage\\8&
\hspace*{0.5cm}earnings\_regular = 8 * hourly\_wage * num\_days&B\\&
\hspace*{0.5cm}earnings\_overtime = overtime\_hours * overtime\_wage * num\_days\\&
\hspace*{0.5cm}total\_earnings = earnings\_regular + earnings\_overtime\\&
\hspace*{0.5cm}return total\_earnings\\&
total\_earnings = calculate\_earnings()\\&
print(total\_earnings)\\&
\texttt{\textasciigrave\textasciigrave\textasciigrave}\\&
\texttt{\textasciigrave\textasciigrave\textasciigrave} output\\&
990.0\\&
\texttt{\textasciigrave\textasciigrave\textasciigrave}\\&
Tina makes \fcolorbox{red}{white}{$990.0$} dollars.
&\\
\hline
& \texttt{\textasciigrave\textasciigrave\textasciigrave}python\\&
def calculate\_pay():\\&
\hspace*{0.5cm}\textcolor{grayishred}{"""Tina makes \$18.00 an hour. If she works more than 8 hours per shift, she is eligible for overtime, which is paid by your hourly wage + 1/2 your hourly wage. If she works 10 hours every day for 5 days, how much money does she make?"""}\\&
\hspace*{0.5cm}hourly\_wage = 18.00\\&
\hspace*{0.5cm}hours\_per\_shift = 10\\&
\hspace*{0.5cm}num\_shifts = 5\\&
\hspace*{0.5cm}overtime\_hours = hours\_per\_shift - 8\\9&
\hspace*{0.5cm}overtime\_wage = hourly\_wage + 0.5 * hourly\_wage&D\\&
\hspace*{0.5cm}pay = (8 * hourly\_wage + overtime\_hours * overtime\_wage) * num\_shifts\\&
\hspace*{0.5cm}return pay\\&
total\_pay = calculate\_pay()\\&
print(total\_pay)\\&
\texttt{\textasciigrave\textasciigrave\textasciigrave}\\&
\texttt{\textasciigrave\textasciigrave\textasciigrave} output\\&
990.0\\&
\texttt{\textasciigrave\textasciigrave\textasciigrave}\\&
Tina makes \fcolorbox{red}{white}{$990.0$} dollars.
&\\
\hline
& \texttt{\textasciigrave\textasciigrave\textasciigrave}python\\&
def total\_earnings():\\&
\hspace*{0.5cm}\textcolor{grayishred}{"""Tina makes \$18.00 an hour. If she works more than 8 hours per shift, she is eligible for overtime, which is paid by your hourly wage + 1/2 your hourly wage. If she works 10 hours every day for 5 days, how much money does she make?"""}\\&
\hspace*{0.5cm}hourly\_wage = 18.00\\&
\hspace*{0.5cm}hours\_per\_shift = 10\\&
\hspace*{0.5cm}overtime\_hours = hours\_per\_shift - 8\\&
\hspace*{0.5cm}num\_shifts = 5\\10&
\hspace*{0.5cm}overtime\_wage = hourly\_wage + 0.5 * hourly\_wage&B\\&
\hspace*{0.5cm}regular\_earnings = 8 * hourly\_wage * num\_shifts\\&
\hspace*{0.5cm}overtime\_earnings = overtime\_hours * overtime\_wage * num\_shifts\\&
\hspace*{0.5cm}total\_earnings = regular\_earnings + overtime\_earnings\\&
\hspace*{0.5cm}return total\_earnings\\&
total\_earnings = total\_earnings()\\&
print(total\_earnings)\\&
\texttt{\textasciigrave\textasciigrave\textasciigrave}\\&
\texttt{\textasciigrave\textasciigrave\textasciigrave} output\\&
990.0\\&
\texttt{\textasciigrave\textasciigrave\textasciigrave}\\&
Tina makes \fcolorbox{red}{white}{$990.0$} dollars.
&\\
\hline
& \texttt{\textasciigrave\textasciigrave\textasciigrave}python\\&
def total\_pay():\\&
\hspace*{0.5cm}\textcolor{grayishred}{"""Tina makes \$18.00 an hour. If she works more than 8 hours per shift, she is eligible for overtime, which is paid by your hourly wage + 1/2 your hourly wage. If she works 10 hours every day for 5 days, how much money does she make?"""}\\&
\hspace*{0.5cm}hourly\_wage = 18.00\\&
\hspace*{0.5cm}hours\_per\_shift = 10\\&
\hspace*{0.5cm}num\_shifts = 5\\&
\hspace*{0.5cm}overtime\_hours = hours\_per\_shift - 8\\&
\hspace*{0.5cm}overtime\_wage = hourly\_wage + 0.5 * hourly\_wage\\11&
\hspace*{0.5cm}regular\_pay = hourly\_wage * 8&D\\&
\hspace*{0.5cm}overtime\_pay = overtime\_wage * overtime\_hours\\&
\hspace*{0.5cm}total\_pay = (regular\_pay + overtime\_pay) * num\_shifts\\&
\hspace*{0.5cm}return total\_pay\\&
total\_pay = total\_pay()\\&
print(total\_pay)\\&
\texttt{\textasciigrave\textasciigrave\textasciigrave}\\&
\texttt{\textasciigrave\textasciigrave\textasciigrave} output\\&
990.0\\&
\texttt{\textasciigrave\textasciigrave\textasciigrave}\\&
Tina makes \fcolorbox{red}{white}{$990.0$} dollars.
&\\
\hline
& \texttt{\textasciigrave\textasciigrave\textasciigrave}python\\&
def calculate\_earnings():\\&
\hspace*{0.5cm}\textcolor{grayishred}{"""Tina makes \$18.00 an hour. If she works more than 8 hours per shift, she is eligible for overtime, which is paid by your hourly wage + 1/2 your hourly wage. If she works 10 hours every day for 5 days, how much money does she make?"""}\\&
\hspace*{0.5cm}hourly\_wage = 18.00\\&
\hspace*{0.5cm}hours\_per\_shift = 10\\&
\hspace*{0.5cm}num\_shifts = 5\\&
\hspace*{0.5cm}overtime\_hours = hours\_per\_shift - 8\\12&
\hspace*{0.5cm}overtime\_wage = hourly\_wage + 0.5 * hourly\_wage&D\\&
\hspace*{0.5cm}earnings = (8 * hourly\_wage) + (overtime\_hours * overtime\_wage)\\&
\hspace*{0.5cm}return earnings * num\_shifts\\&
total\_earnings = calculate\_earnings()\\&
print(total\_earnings)\\&
\texttt{\textasciigrave\textasciigrave\textasciigrave}\\&
\texttt{\textasciigrave\textasciigrave\textasciigrave} output\\&
990.0\\&
\texttt{\textasciigrave\textasciigrave\textasciigrave}\\&
Tina makes \fcolorbox{red}{white}{$990.0$} dollars."
&\\
\hline
& \texttt{\textasciigrave\textasciigrave\textasciigrave}python\\&
def calculate\_salary():\\&
\hspace*{0.5cm}\textcolor{grayishred}{"""Tina makes \$18.00 an hour. If she works more than 8 hours per shift, she is eligible for overtime, which is paid by your hourly wage + 1/2 your hourly wage. If she works 10 hours every day for 5 days, how much money does she make?"""}\\&
\hspace*{0.5cm}hourly\_wage = 18.00\\&
\hspace*{0.5cm}overtime\_wage = hourly\_wage + 0.5 * hourly\_wage\\&
\hspace*{0.5cm}hours\_per\_shift = 10\\&
\hspace*{0.5cm}num\_shifts = 5\\13&
\hspace*{0.5cm}overtime\_hours = hours\_per\_shift - 8 if hours\_per\_shift > 8 else 0&J\\&
\hspace*{0.5cm}salary = (8 * hourly\_wage + overtime\_hours * overtime\_wage) * num\_shifts\\&
\hspace*{0.5cm}return salary\\&
salary = calculate\_salary()\\&
print(salary)\\&
\texttt{\textasciigrave\textasciigrave\textasciigrave}\\&
\texttt{\textasciigrave\textasciigrave\textasciigrave} output\\&
990.0\\&
\texttt{\textasciigrave\textasciigrave\textasciigrave}\\&
Tina makes \fcolorbox{red}{white}{$990.0$} dollars.
&\\
\hline
& \texttt{\textasciigrave\textasciigrave\textasciigrave}python\\&
def calculate\_pay():\\&
\hspace*{0.5cm}\textcolor{grayishred}{"""Tina makes \$18.00 an hour. If she works more than 8 hours per shift, she is eligible for overtime, which is paid by your hourly wage + 1/2 your hourly wage. If she works 10 hours every day for 5 days, how much money does she make?"""}\\&
\hspace*{0.5cm}hourly\_wage = 18.00\\&
\hspace*{0.5cm}overtime\_wage = hourly\_wage + 0.5 * hourly\_wage\\&
\hspace*{0.5cm}hours\_per\_shift = 10\\&
\hspace*{0.5cm}num\_shifts = 5\\&
\hspace*{0.5cm}overtime\_hours = hours\_per\_shift - 8\\14&
\hspace*{0.5cm}regular\_hours = 8&D\\&
\hspace*{0.5cm}total\_pay = (regular\_hours * hourly\_wage + overtime\_hours * overtime\_wage) * num\_shifts\\&
\hspace*{0.5cm}return total\_pay\\&
\\&
total\_pay = calculate\_pay()\\&
print(total\_pay)\\&
\texttt{\textasciigrave\textasciigrave\textasciigrave}\\&
\texttt{\textasciigrave\textasciigrave\textasciigrave} output\\&
990.0\\&
\texttt{\textasciigrave\textasciigrave\textasciigrave}\\&
Tina makes \fcolorbox{red}{white}{$990.0$} dollars.
&\\
\hline
& \texttt{\textasciigrave\textasciigrave\textasciigrave}python\\&
def total\_earnings():\\&
\hspace*{0.5cm}\textcolor{grayishred}{"""Tina makes \$18.00 an hour. If she works more than 8 hours per shift, she is eligible for overtime, which is paid by your hourly wage + 1/2 your hourly wage. If she works 10 hours every day for 5 days, how much money does she make?"""}\\&
\hspace*{0.5cm}hourly\_wage = 18.00\\&
\hspace*{0.5cm}hours\_per\_day = 10\\&
\hspace*{0.5cm}num\_days = 5\\&
\hspace*{0.5cm}overtime\_hours = hours\_per\_day - 8 if hours\_per\_day > 8 else 0\\15&
\hspace*{0.5cm}overtime\_wage = hourly\_wage + 0.5 * hourly\_wage&H\\&
\hspace*{0.5cm}regular\_earnings = hourly\_wage * 8 * num\_days\\&
\hspace*{0.5cm}overtime\_earnings = overtime\_wage * overtime\_hours * num\_days\\&
\hspace*{0.5cm}total\_earnings = regular\_earnings + overtime\_earnings\\&
\hspace*{0.5cm}return total\_earnings\\&
total\_earnings = total\_earnings()\\&
print(total\_earnings)\\&
\texttt{\textasciigrave\textasciigrave\textasciigrave}\\&
\texttt{\textasciigrave\textasciigrave\textasciigrave} output\\&
990.0\\&
\texttt{\textasciigrave\textasciigrave\textasciigrave}\\&
Tina makes \fcolorbox{red}{white}{$990.0$} dollars.
&\\
\hline
\label{tab:random select}
\end{longtable}

\clearpage
\onecolumn
\section{Error Analysis: Numerical Perturb to create $P_5$}
\label{sec:Error Analysis}

\begin{xltabular}{\textwidth}{X} 
\caption{Wrong case for $P_5$: Semantic Ambiguity} \\
\hline
\textbf{Problem:} \\ 
Grandma Jones baked \textcolor{red}{5} apple pies for the fireman's luncheon.  She cut each pie into 8 pieces and set the \textcolor{red}{five} pies out on the buffet table for the guests to serve themselves.  At the end of the evening, after the guests had taken and eaten their pieces of pie, there were 14 pieces of pie remaining.  How many pieces were taken by the guests? \\
\textbf{Completion:} \\ 
\texttt{\textasciigrave\textasciigrave\textasciigrave}python \\
def pie\_pieces\_taken():\\
\hspace*{0.5cm}\textcolor{grayishred}{"""Grandma Jones baked 5 apple pies for the fireman's luncheon.  She cut each pie into 8 pieces and set the five pies out on the buffet table for the guests to serve themselves.  At the end of the evening, after the guests had taken and eaten their pieces of pie, there were 14 pieces of pie remaining.  How many pieces were taken by the guests?"""}\\
\hspace*{0.5cm}\textcolor{red}{pies\_initial = 5}\newline  
\hspace*{0.5cm}pieces\_per\_pie = 8\newline  
\hspace*{0.5cm}total\_pie\_pieces = pies\_initial * pieces\_per\_pie\newline   
\hspace*{0.5cm}remaining\_pie\_pieces = 14\newline   
\hspace*{0.5cm}taken\_pie\_pieces = total\_pie\_pieces - remaining\_pie\_pieces\newline   
\hspace*{0.5cm}return taken\_pie\_pieces\newline \newline 
taken\_pie\_pieces = pie\_pieces\_taken()\newline 
print(taken\_pie\_pieces)\newline 
\texttt{\textasciigrave\textasciigrave\textasciigrave}\newline 
\texttt{\textasciigrave\textasciigrave\textasciigrave}output\newline 
26\newline 
\texttt{\textasciigrave\textasciigrave\textasciigrave}\newline 
The guests took \fcolorbox{red}{white}{$26$} pieces of pie. \\ 
\\
\textbf{Perturbed Problem:} \\ 
Grandma Jones baked \textcolor{red}{10} apple pies for the fireman's luncheon.  She cut each pie into 13 pieces and set the \textcolor{red}{five} pies out on the buffet table for the guests to serve themselves.  At the end of the evening, after the guests had taken and eaten their pieces of pie, there were 19 pieces of pie remaining.  How many pieces were taken by the guests? \\
\textbf{Perturbed Completion:} \\ 
\texttt{\textasciigrave\textasciigrave\textasciigrave}python \\
def pie\_pieces\_taken():\\
\hspace*{0.5cm}\textcolor{grayishred}{"""Grandma Jones baked 10 apple pies for the fireman's luncheon.  She cut each pie into 13 pieces and set the five pies out on the buffet table for the guests to serve themselves.  At the end of the evening, after the guests had taken and eaten their pieces of pie, there were 19 pieces of pie remaining.  How many pieces were taken by the guests?"""}\\
\hspace*{0.5cm}\textcolor{red}{pies\_initial = 10}\newline  
\hspace*{0.5cm}pieces\_per\_pie = 13\newline  
\hspace*{0.5cm}total\_pie\_pieces = pies\_initial * pieces\_per\_pie\newline   
\hspace*{0.5cm}remaining\_pie\_pieces = 19\newline   
\hspace*{0.5cm}taken\_pie\_pieces = total\_pie\_pieces - remaining\_pie\_pieces\newline   
\hspace*{0.5cm}return taken\_pie\_pieces\newline \newline 
taken\_pie\_pieces = pie\_pieces\_taken()\newline 
print(taken\_pie\_pieces)\newline 
\texttt{\textasciigrave\textasciigrave\textasciigrave}\newline 
\texttt{\textasciigrave\textasciigrave\textasciigrave}output\newline 
111\newline 
\texttt{\textasciigrave\textasciigrave\textasciigrave}\newline 
The guests took \fcolorbox{red}{white}{$26$} pieces of pie. \\ 
\hline
\end{xltabular}

\begin{xltabular}{\textwidth}{X} 
\caption{Wrong case for $P_5$: Illogical Numerical Perturb} \\
\hline
\textbf{Problem:} \\ 
A mechanic charges different rates to repair the tires of trucks and cars.  For each truck tire that is repaired, the mechanic will charge \$60 and for each car tire that is repaired, the mechanic will charge \$40.  On Thursday, the mechanic repairs 6 truck tires and 4 car tires.  On Friday, the mechanic repairs 12 car tries and doesn't repair any truck tires.  How much more revenue did the mechanic earn on the day with higher revenue? \\
\textbf{Completion:} \\
\texttt{\textasciigrave\textasciigrave\textasciigrave}python \\
def revenue]\_difference():\\
\hspace*{0.5cm}\textcolor{grayishred}{"""A mechanic charges different rates to repair the tires of trucks and cars.  For each truck tire that is repaired, the mechanic will charge  \$60 and for each car tire that is repaired, the mechanic will charge \$40.  On Thursday, the mechanic repairs 6 truck tires and 4 car tires.  On Friday, the mechanic repairs 12 car tries and doesn't repair any truck tires.  How much more revenue did the mechanic earn on the day with higher revenue?"""}  \\
\hspace*{0.5cm}truck\_tire\_cost = 60\newline 
\hspace*{0.5cm}car\_tire\_cost = 40\newline 
\hspace*{0.5cm}truck\_tires\_repaired\_thursday = 6\newline 
\hspace*{0.5cm}car\_tires\_repaired\_thursday = 4\newline 
\hspace*{0.5cm}car\_tires\_repaired\_friday = 12\newline 
\hspace*{0.5cm}truck\_tires\_repaired\_friday = 0\newline 
\hspace*{0.5cm}revenue\_thursday = (truck\_tire\_cost * truck\_tires\_repaired\_thursday) + (car\_tire\_cost * car\_tires\_repaired\_thursday)\newline 
\hspace*{0.5cm}revenue\_friday = (truck\_tire\_cost * truck\_tires\_repaired\_friday) + (car\_tire\_cost * car\_tires\_repaired\_friday)\newline
revenue\_difference = revenue\_friday - revenue\_thursday\newline
\hspace*{0.5cm}return revenue\_difference\newline 
revenue\_difference = revenue\_difference()\newline 
print(revenue\_difference)\newline
\texttt{\textasciigrave\textasciigrave\textasciigrave}\newline 
\texttt{\textasciigrave\textasciigrave\textasciigrave}output\newline 
\textcolor{red}{-40}\newline \texttt{\textasciigrave\textasciigrave\textasciigrave}\newline 
The mechanic earned \fcolorbox{red}{white}{\textcolor{red}{$40$}} dollars less on Friday than on Thursday. \\

\\
\textbf{Perturbed Problem:} \\ 
A mechanic charges different rates to repair the tires of trucks and cars.  For each truck tire that is repaired, the mechanic will charge  \$65 and for each car tire that is repaired, the mechanic will charge \$45.  On Thursday, the mechanic repairs 11 truck tires and 9 car tires.  On Friday, the mechanic repairs 17 car tries and doesn't repair any truck tires.  How much more revenue did the mechanic earn on the day with higher revenue? \\
\textbf{Perturbed Completion:} \\
\texttt{\textasciigrave\textasciigrave\textasciigrave}python \\
def revenue]\_difference():\\
\hspace*{0.5cm}\textcolor{grayishred}{"""A mechanic charges different rates to repair the tires of trucks and cars.  For each truck tire that is repaired, the mechanic will charge  \$65 and for each car tire that is repaired, the mechanic will charge \$45.  On Thursday, the mechanic repairs 11 truck tires and 9 car tires.  On Friday, the mechanic repairs 17 car tries and doesn't repair any truck tires.  How much more revenue did the mechanic earn on the day with higher revenue?"""}  \\
\hspace*{0.5cm}truck\_tire\_cost = 65\newline 
\hspace*{0.5cm}car\_tire\_cost = 45\newline 
\hspace*{0.5cm}truck\_tires\_repaired\_thursday = 11\newline 
\hspace*{0.5cm}car\_tires\_repaired\_thursday = 9\newline 
\hspace*{0.5cm}car\_tires\_repaired\_friday = 17\newline 
\hspace*{0.5cm}truck\_tires\_repaired\_friday = 0\newline 
\hspace*{0.5cm}revenue\_thursday = (truck\_tire\_cost * truck\_tires\_repaired\_thursday) + (car\_tire\_cost * car\_tires\_repaired\_thursday)\newline 
\hspace*{0.5cm}revenue\_friday = (truck\_tire\_cost * truck\_tires\_repaired\_friday) + (car\_tire\_cost * car\_tires\_repaired\_friday)\newline
revenue\_difference = revenue\_friday - revenue\_thursday\newline
\hspace*{0.5cm}return revenue\_difference\newline 
revenue\_difference = revenue\_difference()\newline 
print(revenue\_difference)\newline
\texttt{\textasciigrave\textasciigrave\textasciigrave}\newline 
\texttt{\textasciigrave\textasciigrave\textasciigrave}output\newline 
\textcolor{red}{355}\newline 
\texttt{\textasciigrave\textasciigrave\textasciigrave}\newline 
The mechanic earned \fcolorbox{red}{white}{\textcolor{red}{$-355$}} dollars less on Friday than on Thursday. \\
\hline
\end{xltabular}

\twocolumn

\end{document}